%% file: main.tex
\pdfoutput=1
\documentclass[11pt]{article}

\usepackage[preprint]{acl}

\usepackage{times}
\usepackage{latexsym}
\usepackage{multirow}

\usepackage[T1]{fontenc}

\usepackage[utf8]{inputenc}

\usepackage[table]{xcolor}
\definecolor{HeaderGray}{gray}{0.9}

\usepackage{microtype}

\usepackage{inconsolata}

\usepackage{graphicx}
\usepackage{booktabs}
\usepackage{enumitem}
\usepackage{amsmath}
\usepackage{amssymb}
\usepackage{tabularx}
\usepackage{array}
\usepackage{makecell}
\usepackage{cuted}
\usepackage{placeins}

\definecolor{HeaderBlue}{RGB}{232,240,248}
\definecolor{GroupGray}{RGB}{247,247,247}
\definecolor{OursBlue}{RGB}{238,247,255}
\definecolor{BestGreen}{RGB}{232,246,238}
\definecolor{ReleaseGreen}{RGB}{205,236,219}

\title{Omni-Embed-Mini: Binding Modalities Without Forgetting\\ via Dense Distillation}

\author{\textbf{Mohammed Irfan Kurpath}, \textbf{Jaseel Muhammad Kaithakkodan},
  \textbf{Sahal Shaji Mullappilly} \\
  \textbf{Ivan Laptev}, \textbf{Hisham Cholakkal} \\
  Mohamed Bin Zayed University of Artificial Intelligence (MBZUAI) \\}

\begin{document}
\raggedbottom
\maketitle

\begin{strip}
\centering
\includegraphics[width=\textwidth]{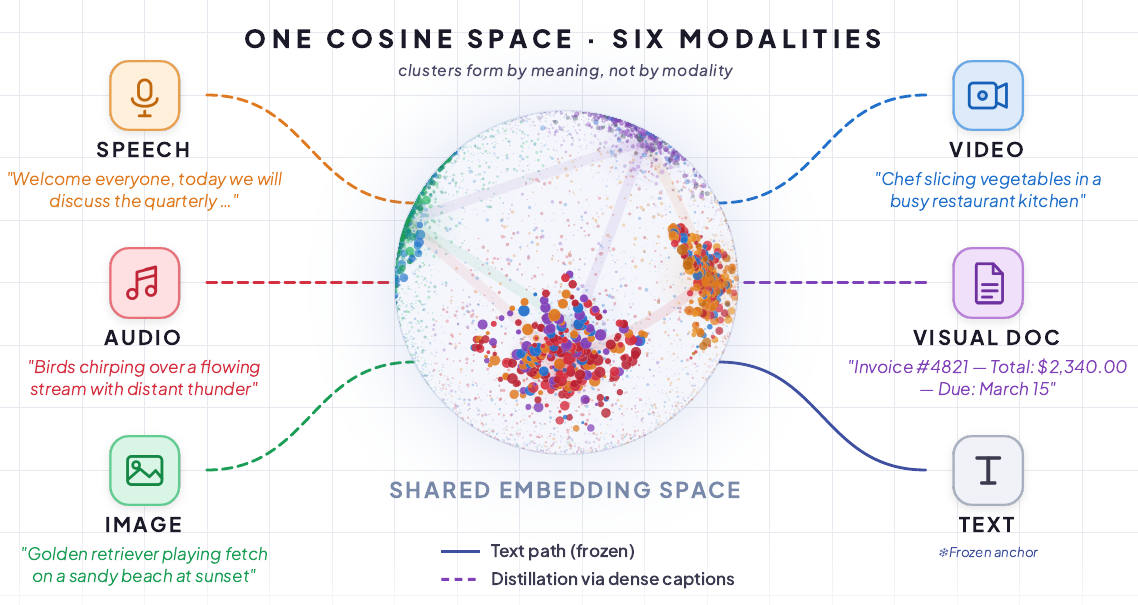}
\captionof{figure}{\textbf{Omni-Embed-Mini overview.} Omni-Embed-Mini maps speech, audio, image, video, visual-document, and text inputs into a single cosine embedding space, where semantically related examples cluster together regardless of modality. The model keeps the text encoder as a frozen anchor and aligns non-text modalities to this shared space through distillation from dense captions.}
\label{fig:teaser}
\end{strip}

\input{sec/0_abstract}

\input{sec/1_intro}
\input{sec/2_related}
\input{sec/3_method}

\input{sec/4_data}
\input{sec/5_experiments}

\input{sec/6_ablations}
\input{sec/7_conclusion}
\input{sec/8_limitations}

\section*{Acknowledgments}
This project is partially supported by research grants from Gates
Foundation (IAAI), and Khalifa bin Zayed Al Awal School, UAE.

\bibliography{references}

\appendix
\input{sec/X_appendix}

\end{document}

%% file: sec/0_abstract.tex
\begin{abstract}
Extending a text embedding model to new modalities typically
\emph{degrades} text retrieval quality, and existing omni-modal
embedders compensate with multi-billion parameters.
We present \textbf{Omni-Embed-Mini}, a 0.9B-parameter model that maps
text, speech, audio, images, video, and visually-rich documents into
a single shared cosine space \emph{without} updating any text-side
parameter.
Our key insight is that the teacher signal requires no separate
embedding model: each media sample is paired with a dense cascaded caption,
and the teacher target is simply the frozen backbone's own embedding
of that caption. Because teacher and student share the same
backbone weights, they inhabit byte-identical geometry, and
lightweight projectors plus phased LoRA adapters on the modality
encoders suffice for alignment.
Training combines a Matryoshka SigLIP contrastive loss with an
online hybrid hard-negative miner whose negatives sharpen as the
encoder improves.
The recipe carries over to a 2.3B variant by swapping in a
native vision-language backbone. Omni-Embed-Mini-0.9B keeps its
text weights bit-identical to the backbone, so training
\emph{cannot} regress text retrieval (49.57 nDCG@10 on MTEB-v2
BEIR-8), while extending it to five additional
modalities, and is ${\sim}$2.7$\times$ to 9.5$\times$ smaller than
every open omni embedder we compare against. The 2.3B variant is
competitive with the closed gemini-embedding-2, edging ahead of it
on the overall-modality average.
Models, code, data and evaluation harness are on our
\href{https://omniembed.cvmbzuai.com}{project page}.
\end{abstract}

%% file: sec/1_intro.tex
\section{Introduction}
\label{sec:intro}

A central goal of recent representation learning is a unified embedding space that binds diverse modalities~\cite{girdhar2023imagebind,zhu2024languagebind}. In practice,
most attempts to jointly contrastive-train text with vision and audio
face two coupled failure modes.
(\textit{i})~\emph{Catastrophic forgetting on the text
side}~\cite{mccloskey1989catastrophic}:
fine-tuning a strong text embedder against media inputs shifts the
text geometry and degrades MTEB performance. Text is still the most critical modality for real-world search, making this degradation highly undesirable. (\textit{ii})~\emph{Modality
bloat}: existing omni embedders such as LCO-Embedding-Omni-3B/7B~\cite{xiao2026scaling},
BidirLM-Omni-2.5B~\cite{boizard2026bidirlm}, NVIDIA's
omni-embed-nemotron-3B~\cite{xu2025omni}, and
e5-omni-3B/7B~\cite{chen2026e5} have scaled to $2.5$B to $8.9$B
parameters in search of enough capacity to absorb every modality,
making them less suited to edge-deployed dynamic retrieval at scale.
We term these two coupled failure modes the \emph{multimodal expansion
dilemma}: under current recipes, every additional modality forces a trade-off, because mitigating text-side catastrophic forgetting has typically come \mbox{with much larger models.}

We argue that text should not be a co-learner; it should be an
\textit{immovable anchor}. BLIP-2~\cite{li2023blip} bridges a frozen image encoder and a frozen
LLM with a small Q-Former, and the first stage of
LLaVA~\cite{liu2023visual} aligns vision to a frozen LLM through a
lightweight projector alone. We
translate this principle to retrieval, drawing on the same logic
\citet{reimers2020making} used to align a new language to a frozen
English teacher's sentence space: each media sample's teacher target is the
EOS-pooled embedding of a dense cascaded caption, computed by the
\emph{identical frozen backbone} that processes the student. Both
targets and text queries are therefore produced by byte-identical
weights; a modest projector and phased LoRA on the encoders close
the modality gap. We call the resulting recipe \textbf{Omni-Embed-Mini}.

As a result, Omni-Embed-Mini offers a favourable size-vs-performance trade-off. Despite being ${\sim}$2.7$\times$ to 9.5$\times$ smaller than the open omni-modal embedders listed above, our $0.9$B model leaves its foundational text retrieval path untrained and covers five additional modalities in a single sub-billion-parameter model; on media retrieval it trails these larger baselines, as expected at its size. Our $2.3$B variant is ahead of the closed gemini-embedding-2 on the overall-modality average ($51.39$ against $49.51$), leading on image, video and visual documents while trailing \mbox{on text, speech and audio.}

Our contribution is not a new alignment primitive. We combine dense
captioning, frozen-backbone self-distillation, parameter-efficient
modality adaptation, Matryoshka training and online mining into an
omni-modal retrieval recipe with one specific invariant: the deployed
text inference path is unchanged. Concretely:
(\textit{i})~\textbf{Zero-regression omni embedding:} a $0.9$B model (Table~\ref{tab:params}) scores 49.57 nDCG@10 on MTEB-v2 BEIR-8 while adding five modalities; the text branch's weights are bit-identical to the stock $0.6$B backbone, so training cannot regress it.
(\textit{ii})~\textbf{Cascaded-caption self-distillation through a shared backbone:} teacher and student share weights, so cosine alignment in the native pooled hidden space suffices: no projection head, no separate embedding teacher, fully cacheable targets.
(\textit{iii})~\textbf{Matryoshka SigLIP + online hybrid hard-negative mining + phased LoRA:} a parameter-efficient recipe that carries over to a $2.3$B variant by swapping the backbone for a native vision-language model.
(\textit{iv})~To our knowledge, this is the \textbf{smallest open omni embedder} at the time of writing, covering text, speech, audio, image, video, and visually-rich documents in one cosine space.

%% file: sec/2_related.tex
\section{Related Work}
\label{sec:related}

\begin{table*}[!t]
\centering
\small
\setlength{\tabcolsep}{5pt}
\begin{tabular*}{\textwidth}{@{\extracolsep{\fill}}llrrrr@{}}
\toprule
\textbf{Model} & \textbf{Backbone} & \textbf{Inference} & \textbf{Frozen} & \textbf{Trained} & \textbf{Trained \%} \\
\midrule
\rowcolor{OursBlue}
\textbf{Omni-Embed-Mini-0.9B} & Qwen3-Embedding-0.6B & \textbf{935.35\,M} & 869.98\,M & 68.32\,M & 7.3\% \\
\rowcolor{OursBlue}
\textbf{Omni-Embed-Mini-2.3B} & Qwen3-VL-Embedding-2B & \textbf{2.44\,B} & 2.30\,B & 141.68\,M & 5.8\% \\
\midrule
BidirLM-Omni-2.5B      & Qwen3-1.7B $\times3$ merged   & 2.45\,B & 724\,M & 1.72\,B & 70.4\% \\
omni-embed-nemotron-3B & Qwen2.5-Omni-3B Thinker & 4.70\,B & 4.67 to 4.70\,B & 7 to 30\,M & 0.2 to 0.6\% \\
e5-omni-3B             & Qwen2.5-Omni-3B Thinker & 4.70\,B & 4.58 to 4.70\,B & 7 to 120\,M & 0.2 to 2.5\% \\
LCO-Embedding-Omni-3B  & Qwen2.5-Omni-3B Thinker & 4.70\,B & 4.58\,B & 120\,M & 2.5\% \\
e5-omni-7B             & Qwen2.5-Omni-7B Thinker & 8.93\,B & 8.77 to 8.92\,B & 10 to 161\,M & 0.1 to 1.8\% \\
LCO-Embedding-Omni-7B  & Qwen2.5-Omni-7B Thinker & 8.93\,B & 8.77\,B & 161\,M & 1.8\% \\
\bottomrule
\end{tabular*}
\caption{\textbf{Inference, frozen and trained parameters} across
open omni embedders. \emph{Inference} is the full parameter count at
inference time; \emph{Trained} is what training updates, and
\emph{Trained \%} is trained over inference. For our 0.9B, the
68.32\,M trained parameters are 65.37\,M of projectors and pooler
plus 2.95\,M of $r{=}16$ LoRA adapters on the media encoders'
attention blocks; the adapters are merged into the encoder weights
at export, so they add nothing at inference and the total stays at
935.35\,M; Table~\ref{supp:tab:params-breakdown} gives the
per-component breakdown. The text backbone is never updated.
gemini-embedding-2 is
omitted because its parameter count is not disclosed.}
\label{tab:params}
\end{table*}

\paragraph{Per-modality embedders.}
\emph{Text.}~The MTEB benchmark family~\cite{muennighoff2023mteb,thakur2021beir}
has strong text encoders such as
Qwen3-Embedding~\cite{zhang2025qwen3}, voyage-3-m-exp~\cite{voyage3},
Conan-embedding-v2~\cite{conanv2}, GritLM~\cite{muennighoff2025generative},
inf-retriever~\cite{infly-ai_2025}, and LENS~\cite{lei2025enhancing};
we treat these as text-side reference points, while our
zero-regression claim is measured against our own frozen backbones:
Qwen3-Embedding-0.6B for the 0.9B, and Qwen3-VL-Embedding-2B,
including its vision paths, for the 2.3B.
\emph{Vision--language.}~VLM2Vec-V2~\cite{meng2025vlm2vec},
Qwen3-VL-Embedding~\cite{li2026qwen3},
RzenEmbed~\cite{jian2025rzenembed}, Ops-MM-embedding~\cite{opsmm},
and seed-1.6~\cite{seed16} adapt pretrained vision-language models for
retrieval, usually with contrastive training, but do not handle audio.
\emph{Audio.}~MS-CLAP~\cite{elizalde2023clap} introduced contrastive
language-audio pretraining and LAION-CLAP~\cite{wu2023large} scaled it
with feature fusion and keyword-to-caption augmentation; Whisper~\cite{radford2023robust} (weakly supervised) and
Dasheng~\cite{dinkel2024scaling} (self-supervised) are strong audio
encoders without an aligned text tower.

\paragraph{Omni-modal embedders.}
LCO-Embedding-Omni-3B/7B~\cite{xiao2026scaling},
BidirLM-Omni-2.5B~\cite{boizard2026bidirlm}, NVIDIA's
omni-embed-nemotron-3B~\cite{xu2025omni}, and
e5-omni-3B/7B~\cite{chen2026e5} target a single space
across text, image, audio and, apart from BidirLM-Omni, video, but
are substantially larger than Omni-Embed-Mini-0.9B. Gemini Embedding
2~\cite{shanbhogue2026gemini} is a closed,
API-only native multimodal embedder whose parameter count is not
disclosed; we report it for reference and exclude it from
size-matched comparisons.
The closest methodological contrast is e5-omni, which \emph{calibrates}
a LoRA-tuned shared space: modality-aware temperatures, a debiased
negative curriculum, and batch whitening with a covariance-alignment
loss together reduce mismatches between modalities. We instead anchor
to a space that is already calibrated and never updated, so no
calibration machinery is needed; our hybrid miner shares only their
focus on hard negatives and false-negative suppression.
e5-omni-7B leads us on the overall average (52.93 against 51.39),
while our 2.3B edges ahead of e5-omni-3B (51.39 against 49.28). We lead on video (55.18 against 32.92 and 44.73) at
a half to a quarter of their parameters, and hold a strong text score at 0.9B.
Notably, in our text suite e5-omni \emph{falls} from 47.80 to 45.34
from 3B to 7B, which is consistent with the cost that
a frozen anchor is designed to avoid.
Crucially, absorbing every modality through
joint contrastive updating puts the backbone's native retrieval
behaviour at risk, a critical concern for real-world deployment.
Among the open omni embedders compared here, ours is the only one
that leaves the backbone entirely frozen, text for the 0.9B variant and text plus
vision for the 2.3B, adding new modalities through external modality
encoders and projectors alone, so as to preserve the backbone's
native capability rather than degrade it
(see Fig.~\ref{fig:bars_omni}).

\paragraph{Anchored vs.\ contrastive joint alignment.}
ImageBind~\cite{girdhar2023imagebind} uses \emph{images} as a
binding modality: a frozen OpenCLIP ViT-H image encoder serves as anchor,
and per-modality encoders are trained with symmetric InfoNCE to match
the image embedding of naturally co-occurring pairs (image-audio from
video, image-depth from RGB-D, etc.). Cross-modal alignment emerges
because each modality is independently pulled toward the shared image
space. LanguageBind~\cite{zhu2024languagebind} adopts the same
anchored-alignment structure but substitutes \emph{language} for
images as the binding modality, freezing a text encoder and
LoRA-adapting per-modality encoders against it.
AudioCLIP~\cite{guzhov2022audioclip} takes a different joint-training
approach, extending CLIP with a three-way contrastive loss across
image, text, and audio.
Omni-Embed-Mini shares LanguageBind's choice of text as anchor but
departs from both ImageBind and LanguageBind in four ways:
(\textit{i})~alignment adds \emph{asymmetric distillation} (the frozen
backbone embeds a dense caption as the teacher target) to the
contrastive objective, so each modality is also fitted to a fixed
per-sample target vector rather than to batch-level contrast alone;
(\textit{ii})~pairing data is \emph{manufactured} via dense
captioning rather than requiring natural co-occurrence, removing
ImageBind's reliance on modality-specific paired corpora;
LanguageBind also uses generated captions as contrastive text,
whereas we additionally use them as a distillation target;
(\textit{iii})~the text encoder is not a separate frozen tower but the
\emph{identical backbone} that serves both teacher and student, so
both are produced by the same untrained weights and text retrieval
cannot regress;
(\textit{iv})~because that backbone is a single causal transformer,
multiple modalities can be interleaved into one token sequence and
encoded in a single pass, rather than each modality passing through
its own tower before a late fusion step. Interleaved multi-modal
queries are therefore native to the architecture rather than an added
capability; we do not isolate their contribution experimentally, and
leave a controlled study to future work.
Conceptually, this is the cross-modal analogue of
\citet{reimers2020making}, who aligned a new language to a frozen
English teacher's sentence space via distillation against translated
sentences; we substitute ``language'' with ``modality'' and
``translation'' with ``dense captioning.'' Unlike their separately
trained student, our text tower is frozen outright.

\begin{figure}[!b]
\centering
\includegraphics[width=\columnwidth]{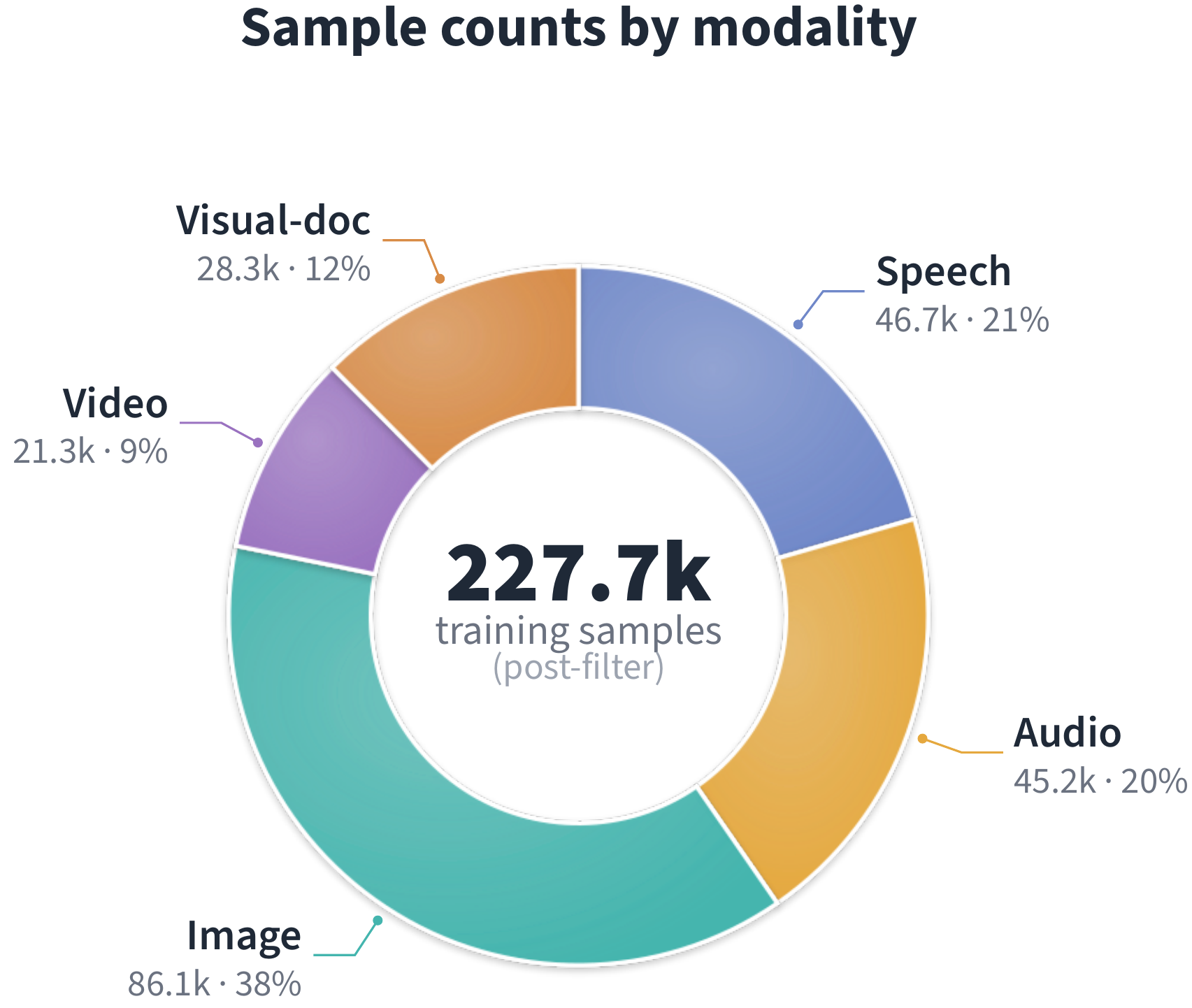}
\caption{\textbf{Training data composition.} Sample-count distribution
across the five non-text modalities in the caption-only slice of our
omni-modal corpus, covering 227k effective rows after caption and
duration filtering.}
\label{fig:data-counts}
\end{figure}

Our recipe builds on several established techniques.
ANCE~\cite{xiong2020approximate} shows that periodically
refreshing the hard-negative pool against the current encoder
substantially improves dense retrievers; our online hybrid miner is
the multimodal-Matryoshka generalisation, with a text index and
per-modality media FAISS-CPU~\cite{johnson2019billion} indices
refreshed on independent cadences.
Matryoshka representation learning~\cite{kusupati2022matryoshka,openai2024embeddingsv3}
enables a single trained embedding to be truncated post-hoc; we
inherit MRL from the backbone and propagate it to all modalities
through the distillation loss. SigLIP~\cite{zhai2023sigmoid} provides
the pair-wise sigmoid contrastive loss, which scores each pair
independently of the batch-level softmax normalisation and so performs
well at small batch size, and LoRA~\cite{hu2021loralowrankadaptationlarge} provides
parameter-efficient adapters on the modality encoders.

%% file: sec/3_method.tex
\section{Method}
\label{sec:method}

\begin{figure}[t]
\centering
\includegraphics[width=\columnwidth]{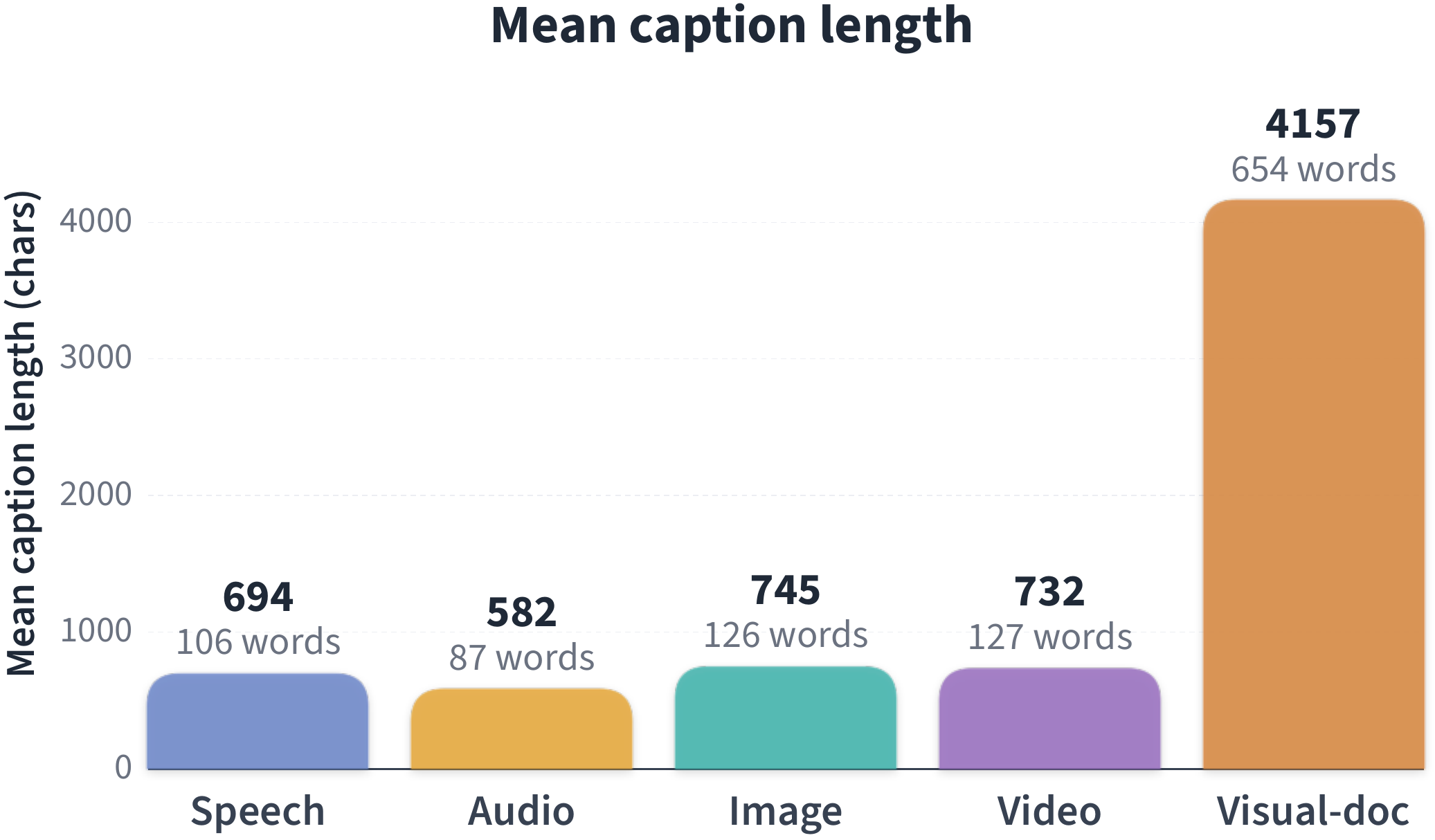}
\caption{\textbf{Cascaded-caption length by modality.} Mean caption
length per modality in the caption-only slice of our omni-modal corpus.
Visual documents are $\sim$6$\times$ longer because the expert-analyst
captioner enumerates layout, headings, tables, and charts.}
\label{fig:data-caption-length}
\end{figure}

\begin{figure*}[!t]
\centering
\includegraphics[width=\textwidth]{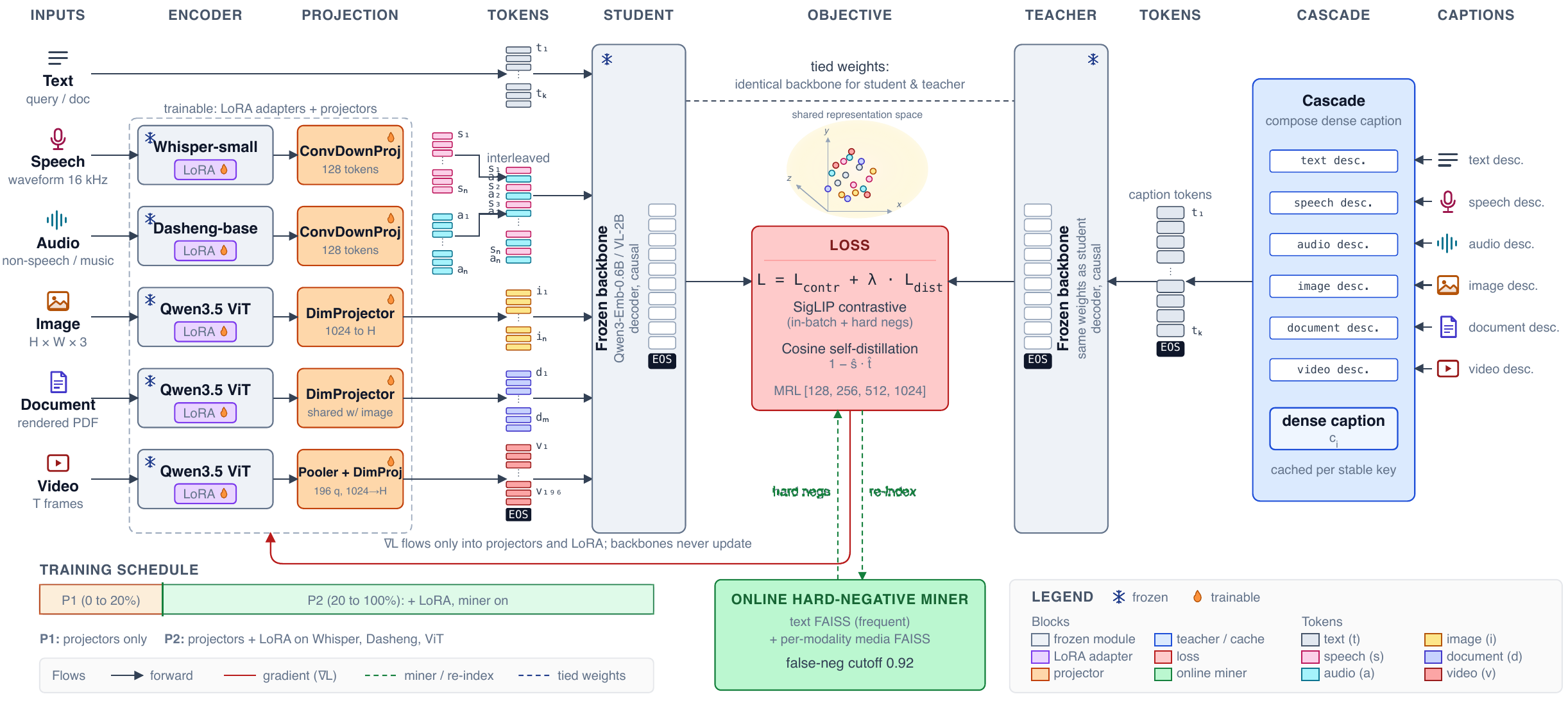}
\caption{\textbf{Omni-Embed-Mini architecture.} Six student inputs flow
through a single \emph{frozen} causal text backbone; a cascaded-caption
teacher mirrors them through the same backbone. Text passes through
unmodified weights, so no text-side parameter is trained. Speech
(Whisper) and audio (Dasheng) each emit 128 projected tokens,
\emph{temporally interleaved} into a 256-token column; image, document
and video share a Qwen3.5 ViT, with video additionally spatially
pooled to 196 tokens. The EOS-pooled hidden state is the embedding.}
\label{fig:arch}
\end{figure*}

\subsection{Asymmetric modality alignment}
\label{sec:method:design}

Two design principles and one training rule govern
Omni-Embed-Mini.
\textbf{(P1) Backbone invariance.} The backbone is never updated:
no full fine-tuning, no LoRA on the backbone. This covers the text
path for the 0.9B variant, and the text and vision paths for the
2.3B, which is built on a vision-language backbone. Those paths carry
no projector and their weights are bit-identical to the stock
backbone, so its calibrated retrieval quality is preserved by
construction.
\textbf{(P2) Cascaded-caption self-distillation.}
Each media sample is paired with a dense caption; the embedding of
that caption, obtained via the \emph{same} frozen backbone, is the
teacher target. Teacher and student share weights, so a cosine
alignment loss in the native pooled hidden space suffices.
\textbf{(T1) Alignment-first, then adaptation.} Phase 1 trains
projectors only; Phase 2 (after 20\% of steps) injects encoder LoRA.
The motivation is that, before the projectors have aligned,
encoder LoRA would adapt to noisy upstream projections; we adopt this
schedule as a design choice and do not ablate the switch point.

\subsection{Architecture}
\label{sec:method:arch}

Fig.~\ref{fig:arch} gives the full data flow for both the student and
the teacher path.

\paragraph{Backbone and pooling.} A decoder-only transformer with
causal attention; an EOS-pooling head extracts the last non-padding
hidden state and L2-normalises it (no projection head). The 0.9B variant uses
Qwen3-Embedding-0.6B~\cite{zhang2025qwen3} (hidden 1024) with a ViT
extracted from Qwen3.5-0.8B~\cite{qwen35}; the 2.3B variant uses
Qwen3-VL-Embedding-2B~\cite{li2026qwen3} (hidden 2048) with native
vision.

\paragraph{Audio path: dual encoder, temporally interleaved.}
Whisper~\cite{radford2023robust} (mel) and
Dasheng~\cite{dinkel2024scaling} (raw waveform) run in parallel; each
emits 128 projected tokens per 30\,s chunk via a 1-D temporal
down-projector with SwiGLU residual. The two streams are
\emph{interleaved in time}, $(w_1, d_1, w_2, d_2, \dots)$, yielding
256 tokens per chunk. Rather than a learned cross-attention bottleneck, temporal
interleaving is intended as a structural inductive bias that encourages the
backbone's causal attention to attend jointly to semantic speech
content and acoustic texture within local context windows, at zero
extra parameter cost.

\paragraph{Vision path.}
\textit{Custom (0.9B):} ViT $\to$ $2{\times}2$ spatial merger $\to$
dimension projector $\to$ backbone hidden size; for video, a spatial
pooler applies cross-attention against a learnable 196-query budget
so video sequence length is independent of frame count.
\textit{Native (2.3B):} the backbone's own visual module handles
image and video; audio is injected via a forward hook on the
embedding layer. Image and video tokens carry multimodal RoPE; audio
tokens are spliced as a contiguous temporal span and reuse text RoPE.
Media embeddings are spliced into modality-tagged placeholder
positions via a one-hot matmul so gradients flow through the
projectors and (Phase 2) encoder LoRA.

\subsection{Cascaded-caption self-distillation}
\label{sec:method:distill}

Each training sample $i$ is paired with a dense caption $c_i$
generated by Qwen3-Omni-30B-A3B-Instruct, with the source dataset's
ground-truth caption pinned in the system prompt as a grounding
prior (\S\ref{supp:datagen}). The teacher embedding $t_i$ is the
L2-normalised EOS-pooled output of the frozen backbone applied to
$c_i$; because the backbone is frozen, $t_i$ is constant across
training and is computed once and cached on disk under a stable
sample id. The student embedding $s_i$ is the same EOS-pool applied
to the token sequence with media spliced in. Because $s_i$ and $t_i$
share the backbone and pooling, they live in identical geometry; no
auxiliary projection head is required.

\subsection{Hybrid Matryoshka contrastive learning}
\label{sec:method:loss}

Let $D=\{128,256,512,1024\,(,\,2048)\}$ denote the set of Matryoshka
dimensions. For each $d\in D$, we truncate the embeddings to their first
$d$ coordinates and re-normalise them.

We compute a SigLIP pair-wise sigmoid loss~\cite{zhai2023sigmoid} over
in-batch and hard-negative pairs, using a learned \mbox{temperature $\tau$ and bias $b$:}
\[
\mathcal{L}_{\mathrm{siglip}}^{(d)}
=
\mathcal{L}_{\mathrm{siglip}}
\bigl(
\hat{s}^{(d)}, \hat{t}^{(d)}; \tau, b
\bigr)
\]

The Matryoshka contrastive loss aggregates the per-dimension losses with
weights proportional to $1/\sqrt{d}$, thereby favouring aggressive
truncations:
\[
w_d
=
\frac{d^{-1/2}}
{\sum_{d'\in D} {d'}^{-1/2}},
\qquad
\mathcal{L}_{\mathrm{contr}}
=
\sum_{d\in D}
w_d\,\mathcal{L}_{\mathrm{siglip}}^{(d)}
\]

We further use cosine self-distillation to align the student and teacher
representations across Matryoshka dimensions:
\[
\mathcal{L}_{\mathrm{dist}}
=
\sum_{d\in D}
w_d\,
\mathbb{E}_{i}
\left[
1-\hat{s}_{i}^{(d)}\cdot \hat{t}_{i}^{(d)}
\right]
\]

The final training objective is
\[
\mathcal{L}
=
\mathcal{L}_{\mathrm{contr}}
+
\lambda\mathcal{L}_{\mathrm{dist}},
\qquad
\lambda = 1
\]

Full equations are given in \S\ref{supp:loss} and implementation
details in \S\ref{supp:hparams}.

\subsection{Online hybrid hard-negative mining}
\label{sec:method:miner}

Hard negatives are supplied by a background subsystem maintaining
FAISS-CPU~\cite{johnson2019billion} inner-product indices (one for
text, one per media modality), in the
spirit of ANCE~\cite{xiong2020approximate}. A lightweight \emph{text} cycle ($\sim$10$\times$/epoch)
re-embeds captions via the text-only path and retains the top-$k$
neighbours per query. A more compute-intensive \emph{media} cycle
($\sim$5$\times$/epoch) re-embeds samples per modality through the
full model into per-modality indices. The miner
shares the live model with the training loop, so each cycle returns
negatives from the \emph{current} representation: as the encoder
improves, mined negatives become harder. Crucially, because the text
branch never receives gradient, the text-side index is perfectly
stable; only the media-side index drifts, and it is refreshed so as
to supply harder negatives as the encoder improves. This stability is intended to guard against
\emph{modality collapse} without ever destabilising text geometry. Concurrency, cache
semantics, and failure handling \mbox{are detailed in \S\ref{supp:miner}.}

%% file: sec/4_data.tex
\section{Training Data: Dense Caption Grounding}
\label{sec:data}

A central claim of this paper is that the success of frozen-backbone
distillation depends critically on the \emph{density} of the
captioning signal. We therefore treat data preparation as a first-class
methodological contribution rather than a logistical detail.

\paragraph{Source corpus.} We train on the caption slice of
our omni-modal caption corpus. From 590{,}858 total rows, filtering to
caption-category rows across the speech, audio, image, video, and
visual-doc configurations yields 242{,}080 single-turn samples;
a subsequent duration filter (removing audio/video clips outside
the supported length range) reduces this to 227{,}704 effective
training rows (Fig.~\ref{fig:data-counts}); caption length varies
sharply by modality (Fig.~\ref{fig:data-caption-length}).
The omni split (cross-modal
multi-turn chat) and chat samples are excluded; their narrow
answers are unsuitable distillation targets.

\paragraph{Cascaded caption pipeline.} Every row is the output of a
single low-temperature ($T{=}0.7$) call to
Qwen3-Omni-30B-A3B-Instruct, with the source dataset's
ground-truth caption interpolated as a ``Reference caption'' inside
a modality-specific system prompt. This \emph{grounds} the cascaded
caption against the source, which may reduce unconstrained
hallucination but does not preclude it. Visual-doc rows use an ``expert document analyst''
prompt with no reference caption (none exists). Full prompts and
worked examples \mbox{are in Appendix \S\ref{supp:datagen}.}

Dense captions are the \emph{single most important data-side
ingredient} of the recipe (as we demonstrate in \S\ref{sec:ablations:dense}): they expose more facets of each sample
(objects, attributes, layout, temporal events, acoustic environment,
OCR text) than short crowd-sourced captions, which provide a weaker,
bag-of-words-like signal. We ablate
this in Table~\ref{tab:dense}, and worked qualitative examples per
modality are in Appendix
Table~\ref{supp:tab:cap-examples-tbl}.

Because every teacher target is machine-generated, caption noise is a
direct risk to the learned geometry. Four properties of the recipe
limit how far it can propagate: a grounded captioning prompt, a
pooled-embedding target rather than token-level supervision, a $0.92$
cosine cutoff in mining, and a frozen text path that no captioner bias
can reach. These contain noise rather than verify its absence; we do
not filter captions for factuality. \S\ref{supp:safeguards} gives the
full argument and its limits.

%% file: sec/5_experiments.tex
\section{Experiments}
\label{sec:exp}

\subsection{Benchmarks}
\label{sec:exp:benchmarks}

\begin{table}[!htbp]
\centering
\small
\setlength{\tabcolsep}{2pt}
\renewcommand{\arraystretch}{0.9}
\begin{tabularx}{\columnwidth}{@{}l*{5}{>{\centering\arraybackslash}X}@{}}
\toprule
\textbf{Style} & \textbf{Sp.} & \textbf{Aud.} & \textbf{Img.} & \textbf{Vid.} & \textbf{Overall} \\
\midrule
Original Captions
& 41.75 & 31.02 & 17.63 & 4.74 & 23.78 \\
\textbf{Dense Captions}
& \textbf{44.43} & \textbf{33.72} & \textbf{26.23} & \textbf{18.80} & \textbf{30.79} \\
\midrule
$\Delta$
& $+2.68$ & $+2.70$ & $+8.60$ & $+14.06$ & $+7.01$ \\
\bottomrule
\end{tabularx}
\caption{\textbf{Dense vs.\ original captions} on the 0.9B mining-ablation checkpoint
(not the final checkpoint, whose scores appear in Fig.~\ref{fig:bars_omni}).
\emph{Original} denotes source-shipped captions, \emph{Dense} our
cascaded captions. Speech/audio are MAEB task means, image/video
MMEB-V2 AvgImg/AvgVid, Overall their mean.}
\label{tab:dense}
\end{table}

\begin{figure*}[!t]
\centering
\includegraphics[width=\textwidth]{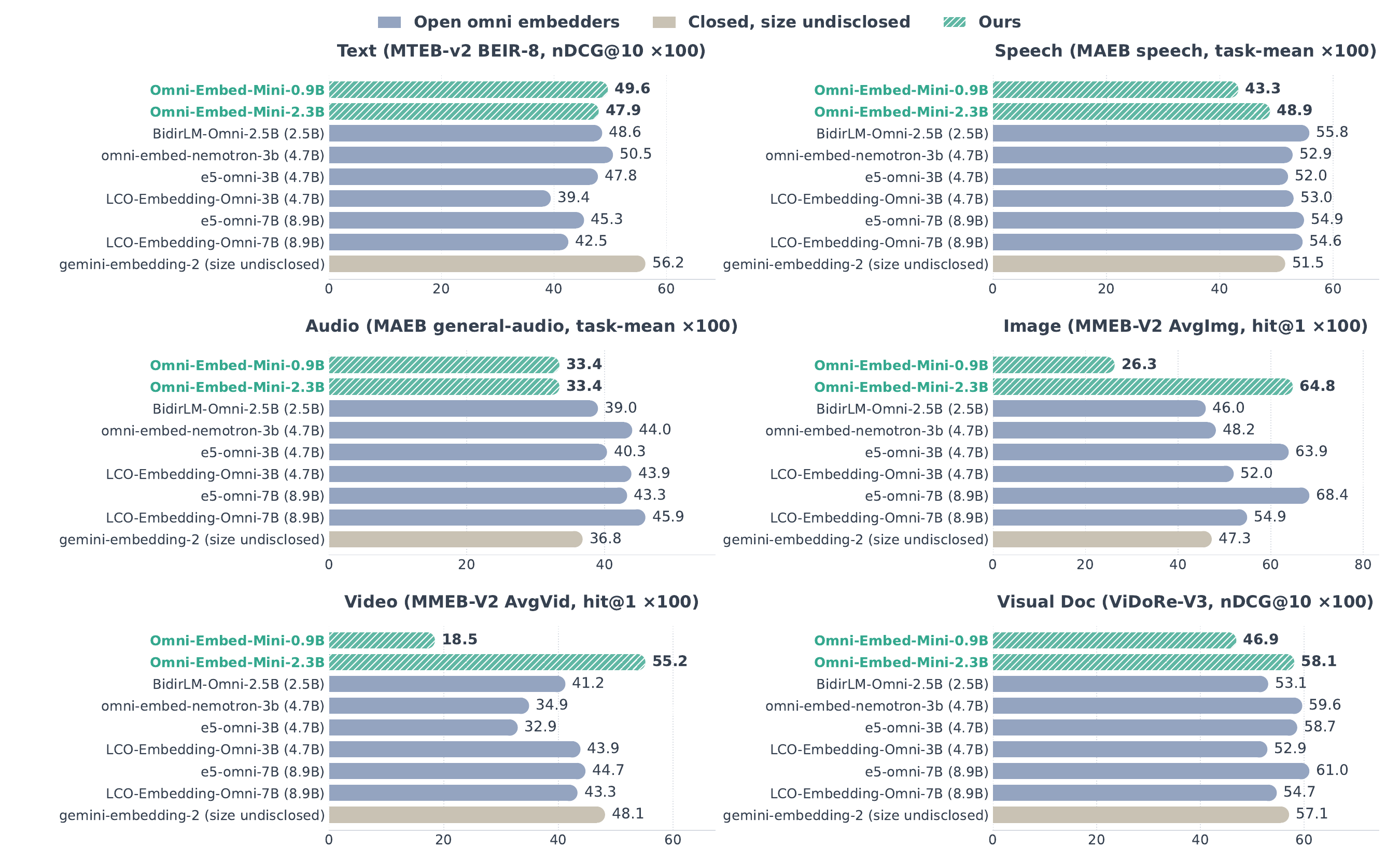}
\caption{\textbf{Per-modality rankings, omni-only.} Same panels as
Appendix Fig.~\ref{supp:fig:bars}, restricted to omni embedders. The
2.3B variant leads the omni block on video and is competitive on image
and visual documents; speech and audio remain our weakest modalities.
gemini-embedding-2 is closed and shown for reference only.}
\label{fig:bars_omni}
\end{figure*}

We evaluate on four community benchmarks; full schemas in
Appendix \S\ref{supp:protocols}.
\textbf{MTEB-v2}~\cite{muennighoff2023mteb,thakur2021beir}: text
retrieval, the 8-task BEIR English subset (ArguAna, CQA-E/P/Pr,
FiQA, NFCorpus, SCIDOCS, SciFact); metric nDCG@10. Role:
\emph{text-side regression test}.
\textbf{MAEB}~\cite{assadi2026maeb}: 22 English audio tasks
(retrieval, classification, clustering, multi-label, pair, reranking,
zero-shot); mixed metrics. Role: cross-modal audio-to-text alignment
plus audio coverage.
\textbf{MMEB-V2}~\cite{meng2025vlm2vec}: 16 English tasks (10
image, 6 video); metric hit@1.
\textbf{ViDoRe-V3}~\cite{loison2026vidore}: 7 English visually-rich
domains; metric nDCG@10.

\paragraph{English-only protocol.} MMEB, MAEB, and ViDoRe are taken
almost in full (16/18, 22/22, 7/7 supported English tasks); MTEB is
reduced to BEIR-8 because the full v2 catalog has 1000+ task-language
pairs. Each benchmark is re-validated by re-running its published
anchor in our harness; full protocols are in
\S\ref{supp:protocols}. Three training sources
share a parent corpus with MMEB-V2 tasks; an identifier-level overlap
audit finds zero shared items \mbox{in every case (\S\ref{supp:overlap}).}

\subsection{Setup}
\label{sec:exp:setup}

\paragraph{Hardware.} 8$\times$ AMD Instinct MI210 (64\,GB HBM2e each), bf16, DDP.

\paragraph{Hyperparameters and cost.} AdamW (wd 0.01, clip 1.0); 2\%
linear warm-up $\to$ cosine; LR $2{\times}10^{-4}$ for audio
projectors, $1{\times}10^{-4}$ for vision projectors and encoder LoRA
(Phase 2 only); LoRA $r{=}16$, $\alpha{=}32$; phase switch at 20\%;
MRL dims $\{128,256,512,1024\}$ ($+2048$ for 2.3B); $\lambda_{\text{dist}}=1$.
Mining ablations are run in a two-epoch setting. Full table in
\S\ref{supp:hparams}.
For video evaluation on MMEB-V2, all omni-style models embed 8
frames as images and mean-pool. The 8 frames are the benchmark's own
pre-extracted frames rather than a sampling choice of ours, so this
keeps comparisons uniform, but it caps \mbox{achievable video performance (\S\ref{supp:limits}).}

\subsection{Main results}
\label{sec:exp:main}

Fig.~\ref{fig:bars_omni} summarises
the headline result restricted to the omni class.
At \textbf{0.9B}, our model is the smallest open omni embedder by
${\sim}$2.7$\times$ relative to the next entry (BidirLM-Omni-2.5B)
and ${\sim}$9.5$\times$ relative to LCO-Embedding-Omni-7B, and it is the only
one among the compared omni embedders whose text branch is untrained
(\S\ref{sec:ablations:zero}). At 49.57 on
\emph{Text} it is second among open omni embedders only to
omni-embed-nemotron-3B (50.48), which is five times larger. On \emph{Speech} and \emph{Audio}
the 0.9B trails the larger omni baselines, the expected cost of
adding these modalities through small projectors rather than by
retraining the backbone. The \textbf{2.3B} variant, built on a frozen
Qwen3-VL-Embedding-2B, reaches an overall-modality average of 51.39:
it leads the omni block on \emph{Video} (55.18) and is competitive on
\emph{Image} (64.80) and \emph{Visual-Doc} (58.10), where e5-omni-7B,
at 3.7$\times$ its size, is the strongest open omni model.
Against gemini-embedding-2, which is closed and of undisclosed size,
our 2.3B leads on the overall-modality average (51.39 to 49.51). The all-class version of
these panels (text-only, VL, audio/CLAP, and omni baselines
together) is in Appendix Fig.~\ref{supp:fig:bars}; per-task tables
are in Appendix~\S\ref{supp:protocols}.

The 0.9B variant scores lower on \emph{Image} (26.29) and \emph{Video} (18.48) because its vision encoder is extracted from a 0.8B backbone with a comparatively small ViT, trained on only ${\sim}$86K image and ${\sim}$36K video rows. The 2.3B variant, with the native Qwen3-VL vision encoder, closes this gap (64.80 image, 55.18 video), which is consistent with vision-encoder capacity, rather than the alignment recipe, limiting the 0.9B.

Routing every modality through a shared space also retains within-modality retrieval without a dedicated training signal. On the image-to-image tasks NIGHTS and FashionIQ (Table~\ref{supp:tab:mmeb-image}), the 2.3B scores 68.00 and 34.50, in line with its frozen Qwen3-VL-Embedding-2B backbone (67.80 and 37.50), and the 0.9B, which is never trained on an image-to-image objective, keeps much of its raw vision encoder's behaviour (53.90 against 59.30 on NIGHTS, 7.40 against 3.80 on \mbox{FashionIQ). See Appendix~\S\ref{supp:emergent}.}

%% file: sec/6_ablations.tex
\section{Ablations and Analysis}
\label{sec:ablations}

Targeted ablations isolate four design choices: frozen backbone, dense captions, hybrid mining, and Matryoshka propagation.

\subsection{Backbone invariance}
\label{sec:ablations:zero}

On MTEB-v2 BEIR-8, Omni-Embed-Mini-0.9B scores \textbf{49.57}.
This is not a tuning artefact: the text-only inference path touches
only the unchanged backbone weights, never the trained projectors or
LoRA adapters, and those weights are bit-identical to the pretrained
Qwen3-Embedding-0.6B. Training consequently cannot move this score,
and independently trained checkpoints agree on all eight tasks to
four decimal places. Per-task numbers are in
Appendix Table~\ref{supp:tab:mteb}. The 2.3B variant inherits the
same guarantee over a wider surface: both its text and its vision
paths are left untrained, so its image and video behaviour is
inherited from the vision-language backbone.

\begin{table}[!t]
\centering
\small
\setlength{\tabcolsep}{1.5pt}
\begin{tabular*}{\columnwidth}{@{\extracolsep{\fill}}lrrrrrr@{}}
\toprule
& \multicolumn{3}{c}{\textbf{0.9B}} & \multicolumn{3}{c}{\textbf{2.3B}} \\
\cmidrule(lr){2-4}\cmidrule(l){5-7}
\textbf{Modality} & \textbf{Frozen} & \textbf{+LoRA} & \textbf{$\Delta$}
                  & \textbf{Frozen} & \textbf{+LoRA} & \textbf{$\Delta$} \\
\midrule
Text       & 49.57 & 46.44 & $-3.13$ & 47.94 & 47.29 & $-0.65$ \\
Speech     & 44.43 & 47.09 & $+2.67$ & 41.55 & 31.02 & $-10.53$ \\
Audio      & 33.72 & 34.83 & $+1.10$ & 33.97 & 25.57 & $-8.40$ \\
Image      & 26.23 & 30.04 & $+3.81$ & 64.80 & 64.34 & $-0.46$ \\
Video      & 18.80 & 24.46 & $+5.66$ & 55.18 & 54.60 & $-0.58$ \\
Vis-Doc    & 46.98 & 51.04 & $+4.06$ & 58.10 & 58.02 & $-0.08$ \\
\bottomrule
\end{tabular*}
\caption{\textbf{Frozen backbone against backbone LoRA.} Both columns
are ablation checkpoints, not the released ones. At 0.9B, unfreezing
the backbone improves every media modality and costs text; at 2.3B it
degrades every modality, and far more on the speech and audio paths
that Phase 2 trains than on the untrained text, \mbox{image and video paths.}}
\label{tab:backbone-lora}
\end{table}

\paragraph{Unfreezing the backbone.} Adding LoRA to the backbone
itself behaves differently at the two scales
(Table~\ref{tab:backbone-lora}). At $0.9$B it is a genuine
trade-off: every media modality improves, by $+1.10$ to $+5.66$,
while text drops by $3.13$ points, which is exactly the regression
the frozen anchor exists to prevent. At $2.3$B the same intervention
degrades every modality, most sharply the two the adapters were
trained for (speech $-10.53$, audio $-8.40$), while text, image and
video shift by under a point; here backbone updates appear to disturb
the alignment that Phase 2 establishes rather than add capacity.
Text improved at neither scale. We therefore keep the backbone
frozen, which preserves text retrieval at both scales and gives the
new modalities a stable target space.

\subsection{Dense vs.\ original captions}
\label{sec:ablations:dense}

Table~\ref{tab:dense} shows the smallest gap is on speech
(+2.68), where source transcripts already carry most of the content;
the largest is on video (+14.06), where short clip captions miss
nearly every temporal event. The same effect motivated the dense recaptioning pipeline
of \citet{li2024if} for VLM pre-training; it is equally
decisive for retrieval.

\subsection{Hybrid mining}
\label{sec:ablations:mining}

\begin{figure}[!tbp]
\centering
\includegraphics[width=\columnwidth]{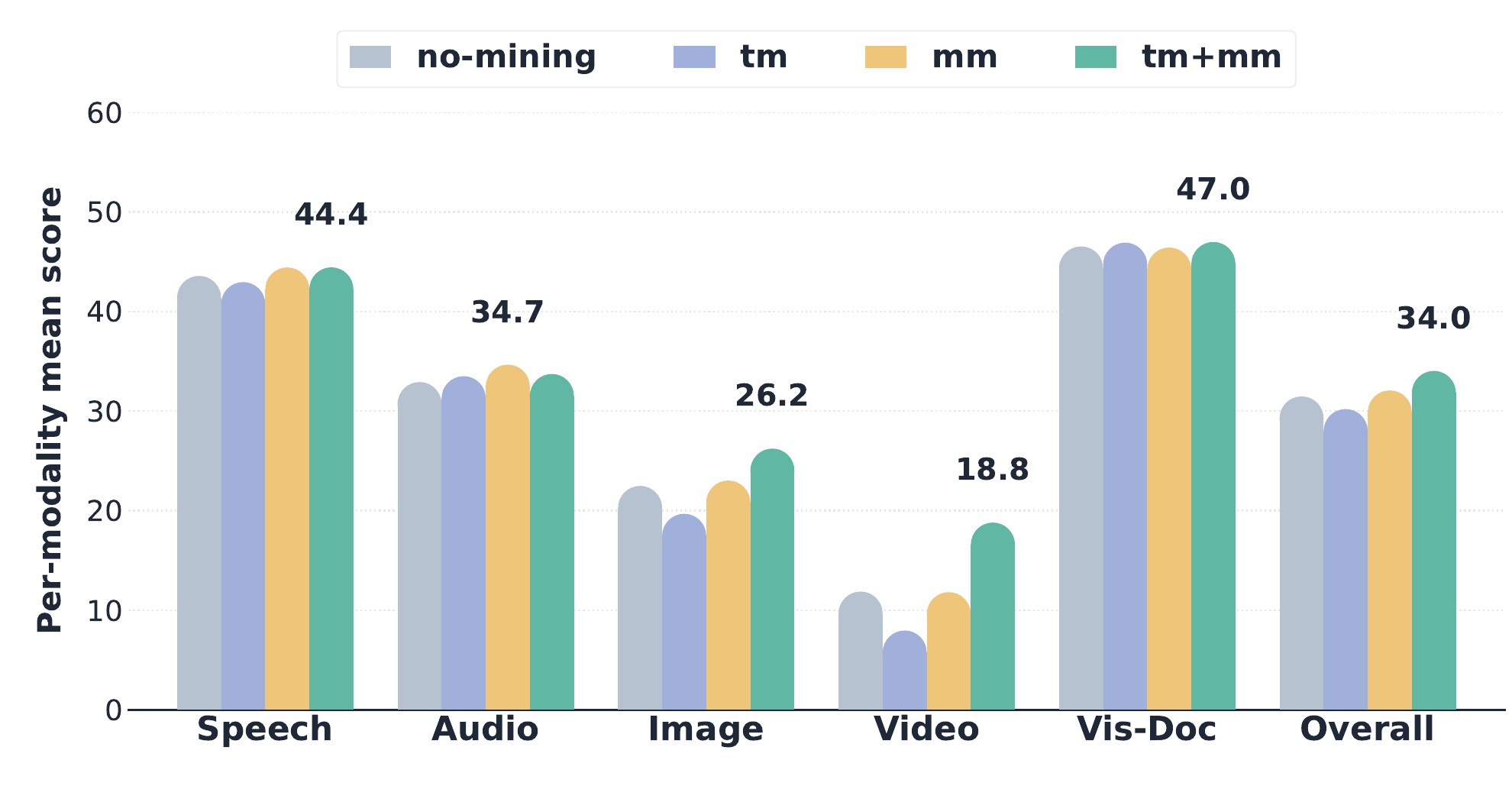}
\caption{\textbf{Mining ablation.} Per-modality means on the 0.9B
recipe across no-mining (in-batch only), text-only, media-only and
combined mining; text is invariant by construction. Combined mining
gives the best overall average and the largest image and video gains;
visual documents are near-insensitive \mbox{(all checkpoints within 0.6\,nDCG@10).}}
\label{fig:mining}
\end{figure}

Fig.~\ref{fig:mining} shows that only combined mining separates
clearly from no mining ($+2.6$ on the mean over the five media
modalities); the gaps among the other three configurations are
comparable to the seed spread reported below. Text-only mining alone slightly hurts on
several modalities (the student sees harder text negatives but no
harder media ones, biasing the projector); media-only gives a small
improvement on most modalities; combining both yields the best overall score,
though per-modality rankings swap on video and on visual documents,
where media-only falls marginally below no-mining. Visual-document
retrieval is in fact the modality least sensitive to mining: the
four configurations agree to within 0.6\,nDCG@10 on ViDoRe-V3
(46.42 to 46.98). What moves that modality is caption quality
rather than negative sampling, as replacing the dense cascaded
captions with the datasets' original captions costs
4.8\,nDCG@10 (46.98 to 42.21). The two miners contribute complementary
pressure: text-mining sharpens the caption-text geometry the teacher
signal targets, while media-mining sharpens the modality-specific
student geometry; dropping either leaves a slack axis along which
the projector can drift. The main recipe uses both miners.

\paragraph{Seed sensitivity.}
Across three random seeds (42, 71 and 1234) of the 10-epoch
no-mining configuration, the model is stable across all five trained
modalities: image scores $27.48 \pm 0.43$, general audio
$33.96 \pm 0.59$, video $19.82 \pm 0.95$, speech $43.10 \pm 1.03$,
and visual-document retrieval $45.48 \pm 1.76$
(mean $\pm$ standard deviation over the three seeds). Text is
identical across all three at 49.57: the text branch carries no
trainable parameters, so every 0.9B checkpoint returns the same
value. Per-seed numbers are
in \S\ref{supp:extra-ablations}.

\subsection{Matryoshka behaviour}
\label{sec:ablations:mrl}

Because MRL slicing acts on the backbone's pooled hidden state with no
learned head, Matryoshka behaviour at every dimension is inherited
from the backbone~\cite{kusupati2022matryoshka,openai2024embeddingsv3}.
Averaged-probe AVG falls only slightly from 1024 to 128 for both
variants, whereas the omni-specific baselines fall away far more
steeply (Fig.~\ref{fig:mrl}). For the 0.9B, a 128-dim truncation
reduces on-disk footprint 8$\times$: a 10M-item omni index in float32
fits in $\sim$5\,GB rather than $\sim$41\,GB, at modest cost in
quality. Further analyses are in \S\ref{supp:extra-ablations}.

%% file: sec/7_conclusion.tex
\section{Conclusion}
\label{sec:conclusion}
We introduced \textbf{Omni-Embed-Mini}, a 0.9B omni-modal retrieval model that binds text, speech, audio, images, video, and visually-rich documents into a unified space by recasting cross-modal alignment as self-distillation through a shared, \emph{frozen} text backbone. By keeping the text encoder fixed and aligning every other modality to its representation, we sidestep the catastrophic forgetting that typically plagues multi-modal fine-tuning and turn omni-modal training into a lightweight projector-plus-LoRA exercise. The model preserves its backbone's text retrieval quality while remaining competitive with open omni embedders ${\sim}$2.7$\times$ to 9.5$\times$ larger, and carries over to a 2.3B variant competitive with baselines 2$\times$ to 4$\times$ its size. Because the backbone is frozen, future text-embedder improvements can in principle be adopted by re-running only the projector and LoRA training, as our backbone swap from the 0.6B text encoder to Qwen3-VL-Embedding-2B illustrates. Open threads include learned mixing of the dual audio-encoder tokens, a Mixture-of-Encoders variant that routes inputs to specialist branches, and replacing per-frame mean-pooling for video (\S\ref{supp:limits}) with a native spatial pooler that better preserves temporal and spatial structure. Beyond architecture, the most pressing community gap is evaluation: existing benchmarks each probe a single modality slice, and none evaluate cross-modal retrieval in arbitrary query$\to$corpus combinations across all six modalities simultaneously.

%% file: sec/8_limitations.tex
\section*{Limitations}
\label{sec:limitations}

Omni-Embed-Mini is a research prototype for omni-modal retrieval and is not validated for high-stakes deployment. Although the model provides efficient alignment across text, speech, audio, image, video, and visually-rich documents, it may still retrieve incorrect, biased, private, or misleading items when used in downstream systems. Therefore, it should not be used as a standalone decision system in sensitive domains without task-specific validation and human oversight.

The model's shared embedding space is constrained by the frozen text backbone used to preserve text-side retrieval performance. As a result, modality-specific phenomena that are difficult to verbalise, such as complex acoustic textures, abstract visual patterns, or fine-grained temporal cues, may not be fully represented. The cascaded-caption self-distillation pipeline also depends on teacher-generated captions, which may contain hallucinations, omissions, or biases that are inherited by the student model. We do not filter captions for factuality, and we do not verify them against their source media; the only automatic check in the pipeline is a 0.92 cosine threshold used to discard likely false negatives during mining. We also did not run a controlled caption-perturbation study, so we cannot quantify how far retrieval quality degrades as caption quality falls. What we can report is that caption quality matters a great deal in aggregate: replacing our dense cascaded captions with the datasets' original captions costs 7.01 points on the mean of speech, audio, image and video (Table~\ref{tab:dense}) and 4.8\,nDCG@10 on visual documents. Because a single captioner produces every target, any systematic bias it carries is applied uniformly across the corpus rather than averaged out, which is the failure mode we would most want to measure. Plausible mitigations we have not implemented include cross-checking captions against a second captioner, filtering on caption-to-media agreement scored by an independent model, and sampling captions for human verification; we regard a caption-robustness study along these lines as the most useful next step for this recipe.

The audio pathway currently uses a fixed temporal interleaving of speech and environmental-audio tokens, which may be suboptimal for inputs dominated by only one audio type. In addition, the compact 0.9B and 2.3B model scales are optimised for efficiency, but may limit performance on complex reasoning-heavy multimodal retrieval tasks.

Finally, the evaluation is limited by the coverage of available benchmarks. Existing datasets do not yet provide a unified any-to-any retrieval benchmark across all six supported modalities, so our results may not generalise to all query-corpus combinations, languages, domains, data qualities, or real-world retrieval workflows.

\section*{Ethical Considerations}
\label{sec:ethics}

This work is intended for research and educational use in omni-modal representation learning and retrieval. Omni-Embed-Mini should not be used for autonomous decision-making in high-stakes domains such as medicine, law, finance, hiring, education, security, or public-sector services without appropriate validation and human supervision.

The main ethical risks include inaccurate retrieval, bias amplification, privacy leakage, and misuse. Since the model aligns multiple modalities into a shared embedding space, errors or biases from training data, teacher-generated captions, and pre-trained encoders may affect retrieval quality across languages, accents, visual appearances, environments, document types, and cultural contexts. Downstream users should conduct safety, fairness, privacy, robustness, and human-in-the-loop evaluations before applying the model \mbox{in sensitive settings.}

We also disclose our use of generative AI tools. Writing-support models were used for grammar, style, and clarity improvements. All technical claims, experimental results, analyses, and conclusions were authored, checked, and verified by the authors; any coding assistance was limited to boilerplate scaffolding, minor refactoring, or formatting support.

%% file: sec/X_appendix.tex
\clearpage

\renewcommand{\topfraction}{0.95}
\renewcommand{\bottomfraction}{0.95}
\renewcommand{\textfraction}{0.05}
\renewcommand{\floatpagefraction}{0.80}
\setcounter{topnumber}{4}
\setcounter{bottomnumber}{4}
\setcounter{totalnumber}{8}
\renewcommand{\dbltopfraction}{0.95}
\renewcommand{\dblfloatpagefraction}{0.80}
\setcounter{dbltopnumber}{4}
\makeatletter
\setlength{\@fptop}{0pt}
\setlength{\@fpsep}{10pt plus 1fil}
\setlength{\@fpbot}{0pt plus 1fil}
\setlength{\@dblfptop}{0pt}
\setlength{\@dblfpsep}{10pt plus 1fil}
\setlength{\@dblfpbot}{0pt plus 1fil}
\makeatother

\section*{Appendix}
\noindent
This Appendix section expands on the main paper with full benchmark
protocols and per-task results for all four benchmarks
(\S\ref{supp:protocols}), the cascaded-caption data-generation
pipeline with system prompts and one qualitative example per modality
(\S\ref{supp:datagen}), per-source dataset attributions and licenses
together with the train/test overlap audit (\S\ref{supp:datasources}),
the mining subsystem (\S\ref{supp:miner}), caption grounding and
noise safeguards (\S\ref{supp:safeguards}), additional ablations
covering Matryoshka truncation, per-seed results and emergent
media-to-media alignment (\S\ref{supp:extra-ablations}), the full loss
equations (\S\ref{supp:loss}), hyperparameters and the per-component
parameter breakdown (\S\ref{supp:hparams}), and \mbox{additional limitations (\S\ref{supp:limits}).}

\section{Detailed benchmark protocols}
\label{supp:protocols}

\begin{figure*}[p]
\centering
\includegraphics[
  width=\textwidth,
  height=0.82\textheight,
  keepaspectratio
]{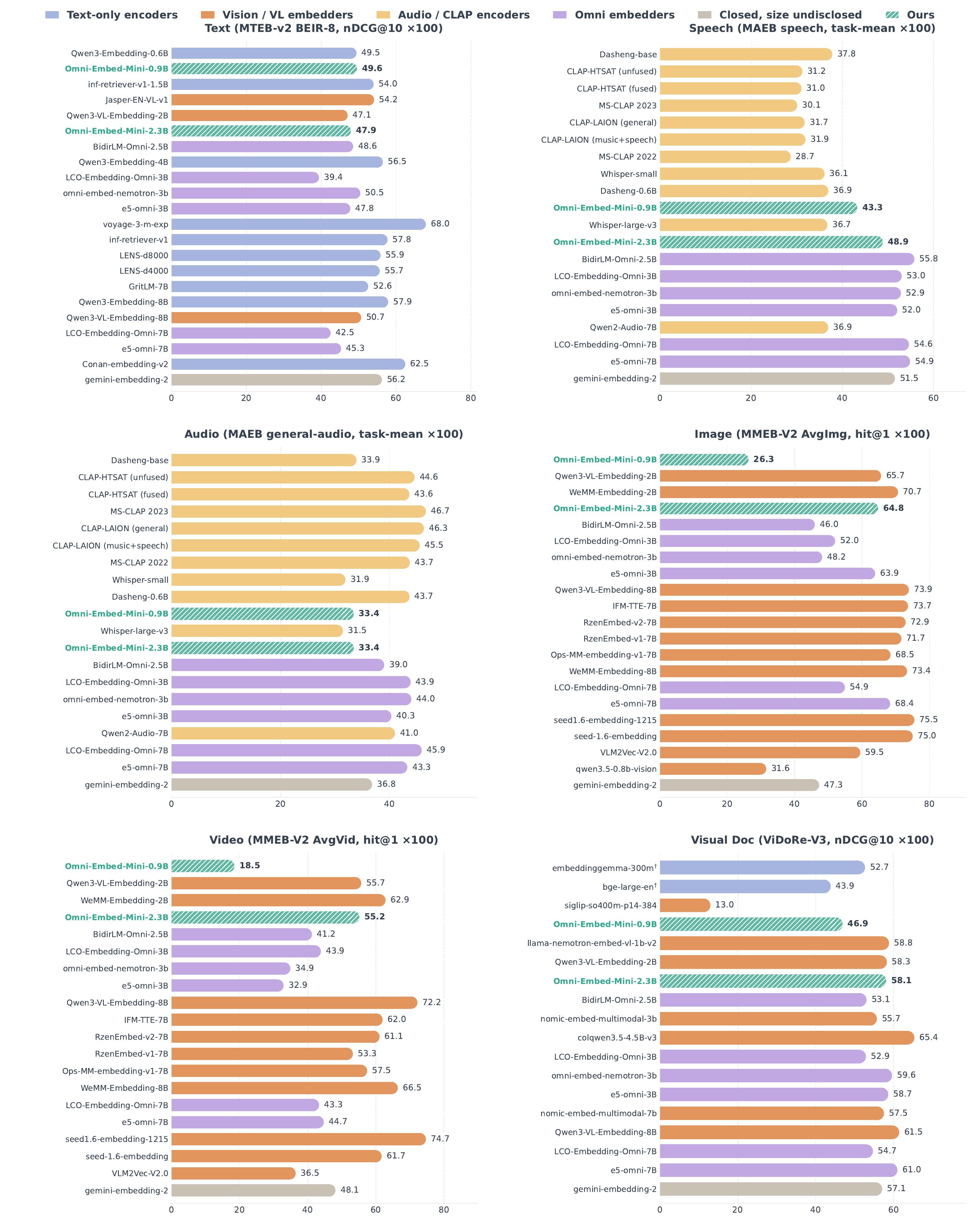}
\caption{\textbf{Per-modality rankings, all classes.} Each panel
shows every model with a score for that modality, ordered by
inference parameter count, smallest first; bars are coloured by class
(text-only, VL/vision, audio/CLAP, omni, ours).
gemini-embedding-2 is closed and of undisclosed size, so it is drawn
in its own colour, placed last, and shown for reference only.
$^\dagger$In the Visual Doc panel, embeddinggemma-300m and
bge-large-en are text retrievers scored on the OCR'd page text alone;
every other model there is scored under the image+text protocol and
sees the page image, so those two bars are not a like-for-like
comparison.
Text is MTEB-v2 BEIR-8 Avg; Speech/Audio are mean across the 12 speech
and 10 general-audio MAEB tasks; Image/Video are MMEB-V2
AvgImg/AvgVid; Visual Doc is ViDoRe-V3 Avg nDCG@10.
Omni-Embed-Mini-0.9B reaches 49.57 MTEB nDCG@10 with text weights
bit-identical to the backbone, so training cannot regress text
retrieval, while extending to five
additional modalities at
${\sim}$2.7$\times$ to 9.5$\times$ fewer parameters than every open omni baseline. The 2.3B
variant additionally inherits Qwen3-VL-Embedding-2B
image/video/visual-doc strengths.}
\label{supp:fig:bars}
\end{figure*}

For each benchmark we briefly describe the source benchmark, the
exact subset of tasks we evaluate, and the metric. We then report
per-task numbers for the established public baselines and our
Omni-Embed-Mini variants. Parameter counts in the leftmost numeric
column are taken from each model's published card; closed-source
models with undisclosed sizes are marked ``?''.

\paragraph{Baselines.} We compare against the following public models.
\emph{Text:} Qwen3-Embedding~\cite{zhang2025qwen3},
voyage-3-m-exp~\cite{voyage3}, Conan-embedding-v2~\cite{conanv2},
inf-retriever~\cite{infly-ai_2025}, LENS~\cite{lei2025enhancing},
GritLM~\cite{muennighoff2025generative},
Jasper~\cite{zhang2025jasperstelladistillationsota},
embeddinggemma~\cite{vera2025embeddinggemma} and
bge-large-en~\cite{xiao2024cpackpackedresourcesgeneral}.
\emph{Vision-language:} Qwen3-VL-Embedding~\cite{li2026qwen3},
VLM2Vec-V2~\cite{meng2025vlm2vec}, RzenEmbed~\cite{jian2025rzenembed},
Ops-MM-embedding~\cite{opsmm}, seed-1.6-embedding~\cite{seed16},
WeMM-Embedding~\cite{zhou2026wemmembeddingwechatmultimodalembedding},
IFM-TTE~\cite{cui2026thinkembedgenerativecontext},
colqwen3.5~\cite{colqwen35v3},
nomic-embed-multimodal~\cite{nomicembedmultimodal2025},
llama-nemotron-embed-vl~\cite{nvidia2025nemotronvl},
SigLIP~\cite{zhai2023sigmoid} and the vision encoder of
Qwen3.5-0.8B~\cite{qwen35}.
\emph{Audio:} MS-CLAP 2022~\cite{elizalde2023clap} and
2023~\cite{elizalde2024naturallanguagesupervisiongeneralpurpose},
LAION-CLAP~\cite{wu2023large}, Qwen2-Audio~\cite{chu2024qwen2}, and
the Whisper~\cite{radford2023robust} and
Dasheng~\cite{dinkel2024scaling} encoders.
\emph{Omni-modal:} LCO-Embedding-Omni~\cite{xiao2026scaling},
BidirLM-Omni~\cite{boizard2026bidirlm},
omni-embed-nemotron~\cite{xu2025omni}, e5-omni~\cite{chen2026e5} and
the closed gemini-embedding-2~\cite{shanbhogue2026gemini}.

\paragraph{Table conventions.} In the per-benchmark tables below, the
\colorbox{OursBlue}{omni block} marks open omni embedders. Within it,
per-column best is in \textbf{bold} and second best
\underline{underlined}, computed over the open omni embedders and our
two \colorbox{ReleaseGreen}{final checkpoints}; the closed
gemini-embedding-2, marked $^\ddagger$, and the
\colorbox{BestGreen}{mining ablation checkpoints} are excluded from
that comparison. gemini-embedding-2 is API-only with an undisclosed
parameter count, so it is reported for reference and excluded from
size-matched comparisons. Indented rows are the hard-negative-mining
ablation checkpoints; the two \colorbox{ReleaseGreen}{highlighted}
rows are our final checkpoints, trained for longer under the best
ablation configuration (text and media mining with dense captions),
and are the models reported in the main text \mbox{and in Fig.~\ref{fig:bars_omni}.}

\subsection{MTEB-v2 (8-task English BEIR subset)}
\label{supp:mteb}

\textbf{What it is.} MTEB-v2~\cite{muennighoff2023mteb} is the second
release of the Massive Text Embedding Benchmark; it modernises the
evaluation API and unifies multilingual (MMTEB) and image (MIEB)
extensions, with the full v2 catalog spanning 1000+ task-language
pairs across 250+ languages.

\textbf{Subset evaluated.} Even at sub-billion parameters, evaluating
the full MTEB-v2 catalog requires days of GPU time, and many
embedding papers report on a curated BEIR subset for the same
reason. We use an 8-task English BEIR subset spanning short and
long queries against short and long corpora: ArguAna,
CQADupstack-English, CQADupstack-Physics, CQADupstack-Programmers,
FiQA-2018, \mbox{NFCorpus, SCIDOCS, SciFact.}

\textbf{Metric.} nDCG@10 ($\times 100$) per task; Avg is the mean
across the 8 tasks. Per-task results are reported in
Table~\ref{supp:tab:mteb}.

\begin{table*}[!tp]
\centering
\setlength{\tabcolsep}{4pt}
\renewcommand{\arraystretch}{1.05}
\resizebox{\textwidth}{!}{%
\begin{tabular}{@{}l r rrrr rrrr r@{}}
\toprule
\textbf{Model} & \textbf{Params}
& \multicolumn{1}{c}{ArgA}
& \multicolumn{1}{c}{CQA-E}
& \multicolumn{1}{c}{CQA-P}
& \multicolumn{1}{c}{CQA-Pr}
& \multicolumn{1}{c}{FiQA}
& \multicolumn{1}{c}{NFC}
& \multicolumn{1}{c}{SCID}
& \multicolumn{1}{c}{SciF}
& \textbf{Avg} \\
\midrule

\multicolumn{11}{@{}l}{\textit{Text embedders}}\\
voyage-3-m-exp           & 6.9B & \textbf{89.79} & \textbf{69.98} & \textbf{73.71} & \textbf{65.87} & \textbf{78.01} & \underline{46.99} & \textbf{34.53} & \underline{85.09} & \textbf{68.00} \\
Conan-embedding-v2       & 1.5B & 77.79 & \underline{57.50} & \underline{60.58} & \underline{60.66} & \underline{71.86} & \textbf{55.74} & 31.63 & 84.27 & \underline{62.50} \\
Qwen3-Embedding-8B       & 7.6B & 76.85 & 57.44 & 58.31 & 53.49 & 64.57 & 41.45 & \underline{32.74} & 78.46 & 57.91 \\
inf-retriever-v1         & 7.1B & \underline{84.86} & 55.96 & 52.35 & 46.64 & 62.35 & 43.66 & 30.78 & \textbf{85.40} & 57.75 \\
Qwen3-Embedding-4B       & 4.0B & 75.64 & 55.00 & 56.58 & 51.17 & 62.65 & 41.10 & 31.44 & 78.33 & 56.49 \\
LENS-d8000               & 7.1B & 76.02 & 55.47 & 53.89 & 51.15 & 61.57 & 40.61 & 28.54 & 79.75 & 55.87 \\
LENS-d4000               & 7.1B & 77.32 & 55.62 & 52.87 & 51.53 & 60.43 & 41.64 & 27.51 & 78.39 & 55.66 \\
inf-retriever-v1-1.5B    & 1.5B & 81.50 & 52.16 & 49.68 & 45.19 & 56.09 & 38.55 & 26.34 & 82.84 & 54.04 \\
GritLM-7B                & 7.2B & 63.17 & 52.10 & 52.85 & 48.29 & 59.91 & 40.86 & 24.41 & 79.13 & 52.59 \\
Qwen3-Embedding-0.6B     & 596M & 69.45 & 51.46 & 50.69 & 46.28 & 47.32 & 36.35 & 24.29 & 69.76 & 49.45 \\

\addlinespace[0.35em]
\multicolumn{11}{@{}l}{\textit{Vision--language embedders}}\\
Jasper-EN-VL-v1          & 1.6B & 65.20 & 54.27 & 55.00 & 50.45 & 60.12 & 43.31 & 24.64 & 80.37 & 54.17 \\
Qwen3-VL-Embedding-8B    & 8.1B & 62.77 & 52.88 & 52.13 & 46.50 & 47.62 & 39.75 & 24.72 & 79.19 & 50.70 \\
Qwen3-VL-Embedding-2B    & 2.1B & 58.46 & 48.27 & 50.66 & 43.44 & 42.30 & 37.05 & 21.79 & 74.73 & 47.09 \\

\midrule
\rowcolor{OursBlue}
\multicolumn{11}{@{}l}{\textit{Omni / audio--language embedders}}\\
\rowcolor{OursBlue}
BidirLM-Omni-2.5B        & 2.5B & 57.04 & 49.91 & 51.67 & 45.93 & 45.60 & \underline{38.67} & 23.32 & 76.31 & 48.56 \\
\rowcolor{OursBlue}
LCO-Embedding-Omni-7B    & 8.9B & 39.09 & 44.47 & 50.59 & 43.18 & 39.21 & 36.35 & 19.97 & 67.42 & 42.54 \\
\rowcolor{OursBlue}
LCO-Embedding-Omni-3B    & 4.7B & 41.39 & 39.51 & 45.67 & 40.14 & 35.47 & 31.86 & 18.21 & 63.13 & 39.42 \\
\rowcolor{OursBlue}
omni-embed-nemotron-3B   & 4.7B & 59.00 & \textbf{55.45} & \textbf{54.64} & \textbf{49.57} & \textbf{53.16} & 36.31 & 21.70 & 74.01 & \textbf{50.48} \\
\rowcolor{OursBlue}
e5-omni-3B               & 4.7B & 52.60 & 46.29 & 50.15 & 41.90 & 51.24 & 37.92 & \textbf{24.46} & \textbf{77.88} & 47.80 \\
\rowcolor{OursBlue}
e5-omni-7B               & 8.9B & 40.17 & 43.20 & 47.81 & 39.91 & \underline{51.29} & \textbf{40.44} & 22.74 & \underline{77.13} & 45.34 \\
\rowcolor{OursBlue}
gemini-embedding-2$^\ddagger$ & ? & 71.61 & 58.16 & 56.47 & 51.62 & 53.44 & 43.74 & 24.82 & 90.10 & 56.24 \\
\rowcolor{ReleaseGreen}
\textbf{Omni-Embed-Mini-0.9B}     & 935M & \textbf{69.54} & \underline{51.62} & 50.77 & \underline{46.17} & 47.47 & 36.83 & \underline{24.38} & 69.80 & \underline{49.57} \\
\rowcolor{ReleaseGreen}
\textbf{Omni-Embed-Mini-2.3B}     & 2.3B & \underline{59.72} & 48.98 & \underline{52.74} & 44.98 & 43.79 & 37.00 & 22.04 & 74.26 & 47.94 \\
\bottomrule
\end{tabular}}
\caption{\textbf{MTEB-v2 BEIR-8 per-task scores.} All eight tasks measure text retrieval ranking quality, scored by nDCG@10 ($\times 100$). \textbf{Task abbreviations:} ArguAna (ArgA), CQADupstack-English (CQA-E), CQADupstack-Physics (CQA-P), CQADupstack-Programmers (CQA-Pr), FiQA-2018 (FiQA), NFCorpus (NFC), SCIDOCS (SCID), SciFact (SciF). Bolding, colour and footnote conventions follow \S\ref{supp:protocols}. Omni-Embed-Mini-0.9B applies no LoRA or projector to its text branch, whose weights are bit-identical to the frozen Qwen3-Embedding-0.6B backbone, so training cannot regress text retrieval.}
\label{supp:tab:mteb}
\end{table*}

\subsection{MAEB (22-task English subset)}
\label{supp:maeb}

\textbf{What it is.} MAEB~\cite{assadi2026maeb} is a unified
MTEB-ecosystem benchmark for audio embeddings that covers speech,
environmental sound, and music with both audio-only and text-query
tasks.

\textbf{Subset evaluated.} We evaluate a 22-task subset comprising the
15 native-English tasks plus 7 language-independent
music/environmental-sound tasks; multilingual source tasks are
restricted to their English split. Tasks split naturally into a
10-task \emph{audio} subset and a 12-task \emph{speech} subset, each
reported in a separate table below. The benchmark task families,
following~\cite{assadi2026maeb}, are: text-to-audio retrieval (T2A),
classification (Cls), clustering (Clu), multi-label tagging (Mlt),
pair-classification (Pair), reranking (Rank) and zero-shot
classification (ZSh).

\textbf{Metrics.} T2A reports HitRate@5; Cls and ZSh report accuracy;
Clu reports V-measure; Mlt reports LRAP; Pair reports max average
precision; Rank reports MAP@1000. Avg is the mean across each model's
scored tasks within the relevant modality subset. Per-task results
are reported in Table~\ref{supp:tab:maeb-audio} (audio) and
Table~\ref{supp:tab:maeb-speech} (speech).

\begin{table*}[!tp]
\centering
\setlength{\tabcolsep}{4pt}
\renewcommand{\arraystretch}{1.05}
\resizebox{\textwidth}{!}{%
\begin{tabular}{@{}l r rrr rrrr r r r r@{}}
\toprule
\textbf{Model} & \textbf{Params}
& \multicolumn{1}{c}{\makecell{Clotho\\\scriptsize HR@5}}
& \multicolumn{1}{c}{\makecell{MACS\\\scriptsize HR@5}}
& \multicolumn{1}{c}{\makecell{US8K\\\scriptsize HR@5}}
& \multicolumn{1}{c}{\makecell{BjOp.\\\scriptsize Acc}}
& \multicolumn{1}{c}{\makecell{Bird\\\scriptsize Acc}}
& \multicolumn{1}{c}{\makecell{GTZAN\\\scriptsize Acc}}
& \multicolumn{1}{c}{\makecell{Mrd.\\\scriptsize Acc}}
& \multicolumn{1}{c}{\makecell{Veh.\\\scriptsize V-meas.}}
& \multicolumn{1}{c}{\makecell{FSD19\\\scriptsize LRAP}}
& \multicolumn{1}{c}{\makecell{GTZ-R\\\scriptsize MAP@1k}}
& \textbf{Avg} \\
\midrule

\multicolumn{13}{@{}l}{\textit{Audio--language models}}\\
Qwen2-Audio-7B          & 7.0B &  0.45 &  1.53 & 0.22 & 97.45 & 37.10 & 93.10 & 61.17 & 5.52 & 32.64 & 80.85 & 41.00 \\

\addlinespace[0.35em]
\multicolumn{13}{@{}l}{\textit{CLAP-style dual encoders}}\\
CLAP-LAION (general)        & 194M & 33.72 & 33.08 & 0.96 & 93.62 & 17.00 & 84.50 & 49.45 & 3.37 & 80.63 & 66.78 & 46.31 \\
CLAP-LAION (music+speech)   & 194M & 32.39 & 30.53 & 0.94 & 91.52 & 16.40 & 83.50 & 48.99 & 4.72 & 80.62 & 65.65 & 45.53 \\
MS-CLAP 2023                & 160M & 41.93 & 27.23 & 0.94 & 91.11 & 17.30 & 78.10 & 52.09 & 2.86 & 79.73 & 75.43 & 46.67 \\
MS-CLAP 2022                & 196M & 58.73 & 41.73 & 0.88 & 93.20 & 13.20 & 58.70 & 29.14 & 7.98 & 70.86 & 62.91 & 43.73 \\
CLAP-HTSAT (unfused)        & 153M & 34.18 & 27.99 & 0.98 & 91.96 & 16.40 & 74.90 & 49.52 & 2.66 & 80.84 & 66.45 & 44.59 \\
CLAP-HTSAT (fused)          & 154M & 33.57 & 29.26 & 0.94 & 92.77 & 18.00 & 67.90 & 47.00 & 4.71 & 81.09 & 61.30 & 43.65 \\

\addlinespace[0.35em]
\multicolumn{13}{@{}l}{\textit{Speech encoders} (no text tower; avg over 7 non-retrieval columns$^\dagger$)}\\
Dasheng-0.6B            & 630M & -- & -- & -- & 84.33 &  8.50 & 76.20 & 53.83 & 5.76 &  7.09 & 69.98 & 43.67$^\dagger$ \\
Dasheng-base            &  86M & -- & -- & -- & 76.73 &  5.10 & 65.40 & 44.50 & 3.22 &  7.08 & 35.56 & 33.94$^\dagger$ \\
Whisper-large-v3        & 1.5B & -- & -- & -- & 64.80 &  5.50 & 45.90 & 30.80 & 1.49 &  7.99 & 63.77 & 31.46$^\dagger$ \\
Whisper-small           & 244M & -- & -- & -- & 67.38 &  7.30 & 47.40 & 27.20 & 0.98 &  9.05 & 64.16 & 31.92$^\dagger$ \\

\midrule
\rowcolor{OursBlue}
\multicolumn{13}{@{}l}{\textit{Omni embedders}}\\
\rowcolor{OursBlue}
LCO-Embedding-Omni-7B   & 8.9B & 33.89 & \textbf{29.77} & \underline{0.92} & \underline{92.79} & \textbf{34.10} & 82.30 & \underline{42.07} & 3.63 & \underline{60.96} & 78.71 & \textbf{45.91} \\
\rowcolor{OursBlue}
LCO-Embedding-Omni-3B   & 4.7B & 32.70 & 22.65 & 0.88 & \textbf{93.22} & 31.60 & 81.00 & 39.54 & 2.18 & 59.67 & 75.37 & 43.88 \\
\rowcolor{OursBlue}
BidirLM-Omni-2.5B       & 2.5B & 12.05 &  8.40 & 0.65 & 90.25 & \underline{33.10} & 74.10 & \textbf{45.35} & \textbf{5.11} & 46.52 & 74.76 & 39.03 \\
\rowcolor{OursBlue}
omni-embed-nemotron-3B   & 4.7B & 33.25 & 25.19 & 0.88 & 88.60 & 28.60 & \underline{82.50} & 32.69 & \underline{4.56} & \textbf{61.60} & 82.12 & \underline{44.00} \\
\rowcolor{OursBlue}
e5-omni-3B              & 4.7B & \underline{39.48} & \underline{27.48} & 0.84 & 76.71 & 24.00 & 78.30 & 30.17 & 3.71 & 37.76 & \underline{84.97} & 40.34 \\
\rowcolor{OursBlue}
e5-omni-7B              & 8.9B & \textbf{42.08} & 26.72 & \textbf{0.96} & 81.81 & 24.50 & \textbf{84.60} & 30.66 & 2.48 & 53.14 & \textbf{85.67} & 43.26 \\
\rowcolor{OursBlue}
gemini-embedding-2$^\ddagger$ & ? & 15.36 & 6.62 & 0.67 & 84.35 & 24.10 & 76.80 & 36.76 & 6.32 & 36.91 & 80.07 & 36.80 \\
\rowcolor{BestGreen}
\multicolumn{13}{@{}l}{\textit{Ours: Omni-Embed-Mini}}\\
\rowcolor{BestGreen}
\quad 0.9B, no mining             & 935M & 13.38 &  6.87 & 0.75 & 79.66 & 15.70 & 66.50 & 28.15 & 2.90 & 39.41 & 75.75 & 32.91 \\
\rowcolor{BestGreen}
\quad 0.9B, text mining           & 935M &  8.16 &  3.82 & 0.67 & 85.60 & 18.40 & 66.10 & 37.22 & 8.19 & 35.30 & 71.56 & 33.50 \\
\rowcolor{BestGreen}
\quad 0.9B, media mining          & 935M & 13.29 &  6.87 & 0.65 & 83.46 & 18.40 & 69.00 & 35.26 & 3.82 & 39.18 & 76.66 & 34.66 \\
\rowcolor{BestGreen}
\quad 0.9B, text + media          & 935M & 12.62 &  7.89 & 0.16 & 81.37 & 14.30 & 66.80 & 30.50 & 8.80 & 39.03 & 75.77 & 33.72 \\
\rowcolor{BestGreen}
\quad 0.9B, t+m, orig.\ captions  & 935M & 18.89 & 10.69 & 0.69 & 74.57 & 12.90 & 50.30 & 37.15 & 6.24 & 34.13 & 64.60 & 31.02 \\
\rowcolor{BestGreen}
\quad 2.3B, text + media          & 2.3B & 26.57 &  9.16 & 0.86 & 77.54 & 16.50 & 66.80 & 26.17 & 1.86 & 39.11 & 75.13 & 33.97 \\
\rowcolor{ReleaseGreen}
\textbf{Omni-Embed-Mini-0.9B}     & 935M &  9.24 &  6.11 & 0.65 & 83.88 & 16.20 & 66.30 & 34.40 & 2.58 & 38.22 & 76.72 & 33.43 \\
\rowcolor{ReleaseGreen}
\textbf{Omni-Embed-Mini-2.3B}     & 2.3B & 17.12 &  8.91 & 0.41 & 84.29 & 18.70 & 65.30 & 32.59 & 0.75 & 34.77 & 71.57 & 33.44 \\
\bottomrule
\end{tabular}}
\caption{MAEB-Audio per-task scores across 10 tasks. \textbf{Task abbreviations:} Clotho-T2A (Clotho), MACS-T2A (MACS), UrbanSound8K-T2A (US8K), BeijingOpera (BjOp.), BirdCLEF (Bird), GTZAN-Genre (GTZAN), MridinghamTonic (Mrd.), VehicleSoundClustering (Veh.), FSD2019Kaggle (FSD19), GTZAN-AudioReranking (GTZ-R). \textbf{Metrics} appear below each task name: HR@5 = HitRate@5, Acc = accuracy, V-meas. = V-measure, LRAP = label ranking avg.\ precision, MAP@1k = mean avg.\ precision at 1000. Bolding, colour and footnote conventions follow \S\ref{supp:protocols}. $^\dagger$Speech-only encoders lack a text tower and are not scored on the three retrieval tasks; their average is over the remaining seven columns.}
\label{supp:tab:maeb-audio}
\end{table*}

\begin{table*}[!tp]
\centering
\setlength{\tabcolsep}{3pt}
\renewcommand{\arraystretch}{1.05}
\resizebox{\textwidth}{!}{%
\begin{tabular}{@{}l r rr rrrr r rrr rr r@{}}
\toprule
\textbf{Model} & \textbf{Params}
& \multicolumn{1}{c}{\makecell{GigaSp\\\scriptsize HR@5}}
& \multicolumn{1}{c}{\makecell{SpSQuAD\\\scriptsize HR@5}}
& \multicolumn{1}{c}{\makecell{CREMA\\\scriptsize Acc}}
& \multicolumn{1}{c}{\makecell{AgeDet\\\scriptsize Acc}}
& \multicolumn{1}{c}{\makecell{IEMOCAP\\\scriptsize Acc}}
& \multicolumn{1}{c}{\makecell{VoxCel\\\scriptsize Acc}}
& \multicolumn{1}{c}{\makecell{CREMA-C\\\scriptsize V-meas.}}
& \multicolumn{1}{c}{\makecell{CREMA-P\\\scriptsize max-AP}}
& \multicolumn{1}{c}{\makecell{NMSQA\\\scriptsize max-AP}}
& \multicolumn{1}{c}{\makecell{VoxPop\\\scriptsize max-AP}}
& \multicolumn{1}{c}{\makecell{Ravd\\\scriptsize ZS Acc}}
& \multicolumn{1}{c}{\makecell{SpCmd\\\scriptsize ZS Acc}}
& \textbf{Avg} \\
\midrule

\multicolumn{15}{@{}l}{\textit{Audio--language models}}\\
Qwen2-Audio-7B          & 7.0B &  0.09 &  1.00 & 73.99 & 17.59 & 92.96 & 29.54 & 32.37 & 68.87 & 48.85 & 52.93 & 14.37 & 10.38 & 36.91 \\

\addlinespace[0.35em]
\multicolumn{15}{@{}l}{\textit{CLAP-style dual encoders}}\\
CLAP-LAION (general)        & 194M &  0.16 &  2.00 & 39.83 & 20.51 & 89.28 & 30.47 & 13.17 & 54.71 & 47.42 & 53.55 & 17.29 & 12.44 & 31.74 \\
CLAP-LAION (music+speech)   & 194M &  0.21 &  2.00 & 40.18 & 16.59 & 93.12 & 33.78 & 14.43 & 56.66 & 46.14 & 53.49 & 17.29 &  9.03 & 31.91 \\
MS-CLAP 2023                & 160M &  0.10 &  0.00 & 37.06 & 15.35 & 85.86 & 26.18 & 10.73 & 57.65 & 51.09 & 52.05 & 15.21 & 10.01 & 30.11 \\
MS-CLAP 2022                & 196M &  0.16 &  2.33 & 27.94 & 14.27 & 89.35 & 25.54 &  5.36 & 53.06 & 50.09 & 52.08 & 13.61 & 10.51 & 28.69 \\
CLAP-HTSAT (unfused)        & 153M &  0.36 &  1.33 & 37.56 & 15.98 & 92.58 & 33.72 & 12.78 & 56.91 & 46.98 & 53.96 & 13.26 &  9.35 & 31.23 \\
CLAP-HTSAT (fused)          & 154M &  0.18 &  2.00 & 38.75 & 15.27 & 93.62 & 29.29 & 10.82 & 54.95 & 47.85 & 53.18 & 14.65 & 11.76 & 31.03 \\

\addlinespace[0.35em]
\multicolumn{15}{@{}l}{\textit{Speech encoders} (no text tower; avg over the 8 non-retrieval / non-zero-shot columns$^\dagger$)}\\
Dasheng-0.6B            & 630M & -- & -- & 39.76 & 20.09 & 57.73 & 16.24 &  8.65 & 55.92 & 46.29 & 50.70 & -- & -- & 36.92$^\dagger$ \\
Dasheng-base            &  86M & -- & -- & 39.04 & 17.22 & 53.91 & 24.10 &  9.13 & 55.94 & 52.38 & 50.70 & -- & -- & 37.80$^\dagger$ \\
Whisper-large-v3        & 1.5B & -- & -- & 41.31 & 17.22 & 50.85 & 16.64 &  9.68 & 57.83 & 48.49 & 51.71 & -- & -- & 36.72$^\dagger$ \\
Whisper-small           & 244M & -- & -- & 37.23 & 19.53 & 52.18 & 13.19 & 10.30 & 58.40 & 45.61 & 52.37 & -- & -- & 36.10$^\dagger$ \\

\midrule
\rowcolor{OursBlue}
\multicolumn{15}{@{}l}{\textit{Omni embedders}}\\
\rowcolor{OursBlue}
LCO-Embedding-Omni-7B   & 8.9B & \textbf{83.29} & 74.00 & \textbf{36.05} & 16.16 & 70.75 & 43.40 & 0.62 & 53.56 & \underline{97.60} & 50.82 & 31.67 & \underline{97.42} & 54.61 \\
\rowcolor{OursBlue}
LCO-Embedding-Omni-3B   & 4.7B & 83.13 & 72.00 & \underline{31.03} & 16.87 & 59.42 & \underline{48.97} & 0.62 & 52.40 & 96.92 & 50.58 & 26.94 & 97.40 & 53.02 \\
\rowcolor{OursBlue}
BidirLM-Omni-2.5B       & 2.5B & 83.05 & 74.33 & 29.92 & 15.04 & \textbf{85.38} & \textbf{49.49} & 0.60 & 53.20 & 95.99 & 51.02 & 34.24 & 97.30 & \textbf{55.80} \\
\rowcolor{OursBlue}
omni-embed-nemotron-3B   & 4.7B & \underline{83.21} & \textbf{81.33} & 21.35 & \textbf{17.12} & 60.45 & 33.05 & \underline{0.66} & 53.68 & \textbf{98.48} & \underline{51.87} & \underline{35.35} & \textbf{97.64} & 52.85 \\
\rowcolor{OursBlue}
e5-omni-3B              & 4.7B & 79.69 & 74.67 & 28.93 & \underline{17.00} & 79.82 & 37.72 & \underline{0.66} & \underline{54.49} & 73.19 & 51.00 & 33.33 & 93.99 & 52.04 \\
\rowcolor{OursBlue}
e5-omni-7B              & 8.9B & 82.80 & \underline{79.67} & 30.69 & 14.85 & \underline{81.26} & 35.89 & \textbf{0.87} & \textbf{55.40} & 90.30 & \textbf{52.13} & \textbf{37.43} & 97.03 & \underline{54.86} \\
\rowcolor{OursBlue}
gemini-embedding-2$^\ddagger$ & ? & 82.34 & 83.33 & 21.74 & 16.84 & 79.25 & 31.92 & 0.55 & 53.40 & 98.05 & 61.56 & 9.65 & 79.82 & 51.54 \\
\rowcolor{BestGreen}
\multicolumn{15}{@{}l}{\textit{Ours: Omni-Embed-Mini}}\\
\rowcolor{BestGreen}
\quad 0.9B, no mining             & 935M & 68.84 & 65.00 & 18.91 & 17.05 & 63.03 & 39.87 & 0.63 & 51.66 & 68.37 & 50.57 & 17.99 & 60.97 & 43.57 \\
\rowcolor{BestGreen}
\quad 0.9B, text mining           & 935M & 66.84 & 58.67 & 19.77 & 14.61 & 58.76 & 38.01 & 0.63 & 52.03 & 69.26 & 50.26 & 18.61 & 67.97 & 42.95 \\
\rowcolor{BestGreen}
\quad 0.9B, media mining          & 935M & 70.19 & 66.00 & 19.35 & 15.91 & 63.74 & 37.26 & 0.58 & 52.02 & 69.66 & 49.79 & 18.68 & 69.83 & 44.42 \\
\rowcolor{BestGreen}
\quad 0.9B, text + media          & 935M & 67.67 & 66.00 & 17.99 & 16.67 & 63.77 & 42.16 & 0.64 & 51.73 & 70.56 & 49.94 & 20.97 & 65.02 & 44.43 \\
\rowcolor{BestGreen}
\quad 0.9B, t+m, orig.\ captions  & 935M & 57.24 & 63.67 & 18.79 & 17.70 & 56.68 & 40.77 & 0.54 & 51.33 & 73.63 & 50.15 & 13.96 & 56.53 & 41.75 \\
\rowcolor{BestGreen}
\quad 2.3B, text + media          & 2.3B & 63.90 & 52.67 & 20.20 & 16.93 & 64.87 & 36.79 & 0.88 & 52.43 & 60.49 & 51.47 & 17.36 & 60.68 & 41.56 \\
\rowcolor{ReleaseGreen}
\textbf{Omni-Embed-Mini-0.9B}     & 935M & 70.42 & 66.33 & 19.30 & 15.07 & 59.10 & 40.71 & 0.57 & 52.22 & 69.13 & 50.41 & 15.56 & 60.53 & 43.28 \\
\rowcolor{ReleaseGreen}
\textbf{Omni-Embed-Mini-2.3B}     & 2.3B & 81.26 & 68.33 & 18.38 & 15.53 & 63.23 & 43.46 & 0.55 & 51.66 & 87.74 & 49.73 & 14.44 & 92.05 & 48.86 \\
\bottomrule
\end{tabular}}
\caption{MAEB Speech per-task scores across 12 tasks. \textbf{Task abbreviations:} GigaSpeechT2A (GigaSp), SpokenSQuADT2A (SpSQuAD), CREMA-D (CREMA), CommonLanguageAgeDetection (AgeDet), IEMOCAPGender (IEMOCAP), VoxCelebSA (VoxCel), CREMA-D-Clustering (CREMA-C), CREMA-D-PairClassification (CREMA-P), NMSQA-PairClassification (NMSQA), VoxPopuli-AccentPairClassification (VoxPop), Ravdess-Zeroshot (Ravd), SpeechCommands-Zeroshot-v0.02 (SpCmd). \textbf{Metrics} appear below each task name: HR@5 = HitRate@5, Acc = accuracy, V-meas. = V-measure, max-AP = pair-classification max average precision, ZS Acc = zero-shot accuracy. Bolding, colour and footnote conventions follow \S\ref{supp:protocols}. $^\dagger$Speech-only encoders lack a text tower and are not scored on the two retrieval tasks or the two zero-shot tasks; their average is over the remaining eight columns.}
\label{supp:tab:maeb-speech}
\end{table*}

\subsection{MMEB-V2 (16-task English subset)}
\label{supp:mmeb}

\textbf{What it is.} MMEB-V2~\cite{meng2025vlm2vec} is a multimodal
embedding benchmark combining image and video tasks under a unified
hit@1 retrieval and classification protocol.

\textbf{Subset evaluated.} We use the 16 English tasks: 10 image
tasks (VOC2007, Country211, OK-VQA, EDIS, MSCOCO-T2I, VisualNews-T2I,
MSCOCO-I2T, VisualNews-I2T, NIGHTS, FashionIQ) and 6 video tasks
(HMDB51, UCF101, MSR-VTT, MSVD, DiDeMo, VATEX). Following the
benchmark task taxonomy, image tasks span zero-shot image
classification (CLS), visual question answering (VQA),
text-to-image (T2I), image-to-text (I2T) and image-to-image (I2I)
retrieval; video tasks span zero-shot action classification (V-CLS)
and text-to-video (V-T2V) retrieval. The MMEB-V2 NExTQA and
EgoSchema video-QA tasks are not included in this subset.

\textbf{Metric.} hit@1 ($\times 100$) per task; AvgImg, AvgVid and
AvgAll are means across the image, video, and combined task sets
respectively. Per-task results are reported in
Table~\ref{supp:tab:mmeb-image} (image tasks) and
Table~\ref{supp:tab:mmeb-video} (video tasks).

\begin{table*}[!tp]
\centering
\setlength{\tabcolsep}{4pt}
\renewcommand{\arraystretch}{1.05}
\resizebox{\textwidth}{!}{%
\begin{tabular}{@{}l r rr r rrr rr rr r@{}}
\toprule
\textbf{Model} & \textbf{Params}
& \multicolumn{1}{c}{\makecell{VOC07\\\scriptsize CLS}}
& \multicolumn{1}{c}{\makecell{Ctry211\\\scriptsize CLS}}
& \multicolumn{1}{c}{\makecell{OKVQA\\\scriptsize VQA}}
& \multicolumn{1}{c}{\makecell{EDIS\\\scriptsize T2I}}
& \multicolumn{1}{c}{\makecell{COCO-T2I\\\scriptsize T2I}}
& \multicolumn{1}{c}{\makecell{VN-T2I\\\scriptsize T2I}}
& \multicolumn{1}{c}{\makecell{COCO-I2T\\\scriptsize I2T}}
& \multicolumn{1}{c}{\makecell{VN-I2T\\\scriptsize I2T}}
& \multicolumn{1}{c}{\makecell{NIGHTS\\\scriptsize I2I}}
& \multicolumn{1}{c}{\makecell{FashIQ\\\scriptsize I2I}}
& \textbf{Avg} \\
\midrule

\multicolumn{13}{@{}l}{\textit{Vision--language embedders}}\\
seed1.6-embedding-1215      & ?    & 89.30 & 46.90 & 74.90 & 96.70 & 81.30 & 83.80 & 77.80 & 83.70 & 70.70 & 50.40 & 75.55 \\
seed-1.6-embedding          & ?    & 91.90 & 47.70 & 74.20 & 91.50 & 78.80 & 83.40 & 77.40 & 84.50 & 72.10 & 49.00 & 75.05 \\
Qwen3-VL-Embedding-8B       & 8.1B & 93.40 & 27.30 & 77.80 & 96.30 & 81.10 & 81.10 & 79.10 & 85.70 & 72.70 & 44.50 & 73.90 \\
IFM-TTE-7B                  & 8.3B & 83.40 & 60.50 & 83.40 & 95.50 & 77.70 & 79.20 & 72.60 & 82.50 & 69.40 & 32.40 & 73.66 \\
WeMM-Embedding-8B           & 8.8B & 94.50 & 32.20 & 73.40 & 94.90 & 81.20 & 81.70 & 78.20 & 84.90 & 69.00 & 43.80 & 73.38 \\
RzenEmbed-v2-7B             & 8.3B & 91.40 & 29.70 & 73.10 & 97.20 & 81.10 & 82.60 & 78.00 & 85.80 & 69.00 & 41.50 & 72.94 \\
RzenEmbed-v1-7B             & 8.3B & 91.60 & 29.00 & 69.60 & 96.50 & 78.60 & 82.60 & 76.00 & 85.80 & 68.60 & 38.70 & 71.70 \\
WeMM-Embedding-2B           & 2.1B & 93.70 & 28.90 & 66.40 & 93.30 & 77.60 & 76.90 & 77.00 & 80.30 & 69.60 & 43.20 & 70.69 \\
Ops-MM-embedding-v1-7B      & 8.3B & 84.80 & 28.50 & 70.60 & 94.40 & 79.30 & 80.10 & 72.10 & 84.30 & 66.20 & 24.30 & 68.46 \\
Qwen3-VL-Embedding-2B       & 2.1B & 84.70 & 20.90 & 65.60 & 88.30 & 75.40 & 70.40 & 71.90 & 74.20 & 67.80 & 37.50 & 65.67 \\
VLM2Vec-V2.0                & 2.2B & 76.80 & 19.50 & 44.00 & 82.90 & 73.80 & 72.30 & 66.40 & 73.70 & 66.20 & 19.30 & 59.49 \\
qwen3.5-0.8b-vision (I2I only) & 100M & -- & -- & -- & -- & -- & -- & -- & -- & 59.30 &  3.80 & 31.55$^\dagger$ \\

\midrule
\rowcolor{OursBlue}
\multicolumn{13}{@{}l}{\textit{Omni embedders}}\\
\rowcolor{OursBlue}
LCO-Embedding-Omni-7B       & 8.9B & 56.40 & \textbf{33.20} & 59.90 & 66.70 & 64.60 & 62.40 & 61.20 & 61.20 & 62.90 & 20.70 & 54.92 \\
\rowcolor{OursBlue}
LCO-Embedding-Omni-3B       & 4.7B & 71.20 & 28.90 & 58.20 & 56.50 & 59.50 & 55.20 & 59.40 & 59.20 & 56.20 & 15.90 & 52.02 \\
\rowcolor{OursBlue}
omni-embed-nemotron-3B      & 4.7B & 53.40 & 23.20 & 16.10 & 82.60 & 60.50 & 61.10 & 52.50 & 59.50 & 64.10 &  8.60 & 48.16 \\
\rowcolor{OursBlue}
BidirLM-Omni-2.5B           & 2.5B & 68.10 &  8.00 & 20.30 & 72.90 & 72.10 & 48.50 & 63.60 & 30.60 & 62.60 & 13.10 & 45.98 \\
\rowcolor{OursBlue}
e5-omni-3B                  & 4.7B & 76.80 & 28.30 & 63.20 & 86.10 & \underline{74.90} & \underline{73.00} & 71.60 & \underline{78.60} & 67.10 & 19.40 & 63.90 \\
\rowcolor{OursBlue}
e5-omni-7B                  & 8.9B & \textbf{84.80} & \underline{32.60} & \textbf{71.00} & \textbf{91.80} & \textbf{76.90} & \textbf{77.30} & \textbf{74.90} & \textbf{81.70} & \underline{67.60} & \underline{25.00} & \textbf{68.36} \\
\rowcolor{OursBlue}
gemini-embedding-2$^\ddagger$ & ? & 67.40 & 13.80 & 17.50 & 87.90 & 74.60 & 82.30 &  2.80 & 45.60 & 64.10 & 16.70 & 47.27 \\
\rowcolor{BestGreen}
\multicolumn{13}{@{}l}{\textit{Ours: Omni-Embed-Mini}}\\
\rowcolor{BestGreen}
\quad 0.9B, no mining             & 935M & 71.30 &  1.50 &  7.00 & 42.00 & 25.60 &  7.60 & 21.10 &  4.70 & 41.50 &  2.50 & 22.48 \\
\rowcolor{BestGreen}
\quad 0.9B, text mining           & 935M & 68.40 &  1.50 &  4.50 & 42.50 & 15.10 &  4.40 & 13.40 &  5.00 & 40.10 &  1.90 & 19.68 \\
\rowcolor{BestGreen}
\quad 0.9B, media mining          & 935M & 61.50 &  1.60 &  4.00 & 40.60 & 22.90 &  8.30 & 23.10 &  8.10 & 55.40 &  4.70 & 23.02 \\
\rowcolor{BestGreen}
\quad 0.9B, text + media          & 935M & 61.40 &  1.00 &  9.60 & 47.70 & 34.40 &  9.60 & 28.90 & 11.50 & 51.00 &  7.20 & 26.23 \\
\rowcolor{BestGreen}
\quad 0.9B, t+m, orig.\ captions  & 935M & 69.20 &  1.40 &  1.00 & 33.60 &  7.50 &  3.20 &  8.70 &  3.90 & 45.30 &  2.50 & 17.63 \\
\rowcolor{BestGreen}
\quad 2.3B, text + media          & 2.3B & 83.80 & 20.90 & 65.00 & 86.70 & 72.70 & 70.90 & 72.30 & 73.20 & 68.00 & 34.50 & 64.80 \\
\rowcolor{ReleaseGreen}
\textbf{Omni-Embed-Mini-0.9B}     & 935M & 60.60 &  1.60 &  8.80 & 46.00 & 29.40 &  8.70 & 35.40 & 11.10 & 53.90 &  7.40 & 26.29 \\
\rowcolor{ReleaseGreen}
\textbf{Omni-Embed-Mini-2.3B}     & 2.3B & \underline{83.80} & 20.90 & \underline{65.00} & \underline{86.70} & 72.70 & 70.90 & \underline{72.30} & 73.20 & \textbf{68.00} & \textbf{34.50} & \underline{64.80} \\
\bottomrule
\end{tabular}}
\caption{MMEB-V2 image per-task hit@1 ($\times 100$). \textbf{Task abbreviations:} VOC2007 (VOC07), Country211 (Ctry211), OK-VQA (OKVQA), EDIS, MSCOCO-T2I (COCO-T2I), VisualNews-T2I (VN-T2I), MSCOCO-I2T (COCO-I2T), VisualNews-I2T (VN-I2T), NIGHTS, FashionIQ (FashIQ). \textbf{Task families} appear below each task name: CLS = zero-shot classification accuracy; VQA = knowledge retrieval hit@1; T2I = text-to-image retrieval hit@1; I2T = image-to-text retrieval hit@1; I2I = image-to-image retrieval hit@1. Bolding, colour and footnote conventions follow \S\ref{supp:protocols}. $^\dagger$qwen3.5-0.8b-vision supports only the two I2I tasks; its average is over those two columns and is not directly comparable to the others.}
\label{supp:tab:mmeb-image}
\end{table*}

\begin{table*}[!tp]
\centering
\setlength{\tabcolsep}{4pt}
\renewcommand{\arraystretch}{1.05}
\resizebox{\textwidth}{!}{%
\begin{tabular}{@{}l r rr rrrr r@{}}
\toprule
\textbf{Model} & \textbf{Params}
& \multicolumn{1}{c}{\makecell{HMDB\\\scriptsize V-CLS}}
& \multicolumn{1}{c}{\makecell{UCF\\\scriptsize V-CLS}}
& \multicolumn{1}{c}{\makecell{MSRVTT\\\scriptsize V-T2V}}
& \multicolumn{1}{c}{\makecell{MSVD\\\scriptsize V-T2V}}
& \multicolumn{1}{c}{\makecell{DiDeMo\\\scriptsize V-T2V}}
& \multicolumn{1}{c}{\makecell{VATEX\\\scriptsize V-T2V}}
& \textbf{Avg} \\
\midrule

\multicolumn{9}{@{}l}{\textit{Vision--language embedders}}\\
seed1.6-embedding-1215  & ?    & 93.50 & 98.70 & 60.20 & 74.48 & 66.43 & 54.87 & 74.70 \\
Qwen3-VL-Embedding-8B   & 8.1B & 83.40 & 95.10 & 58.20 & 75.67 & 66.04 & 54.87 & 72.21 \\
WeMM-Embedding-8B       & 8.8B & 61.50 & 84.20 & 55.30 & 73.43 & 69.22 & 55.05 & 66.45 \\
WeMM-Embedding-2B       & 2.1B & 57.70 & 75.10 & 53.10 & 73.43 & 65.24 & 52.52 & 62.85 \\
IFM-TTE-7B              & 8.3B & 65.40 & 79.60 & 52.70 & 73.13 & 49.70 & 51.45 & 62.00 \\
seed-1.6-embedding      & ?    & 63.90 & 74.20 & 55.30 & 71.34 & 56.67 & 48.77 & 61.70 \\
RzenEmbed-v2-7B         & 8.3B & 63.40 & 75.70 & 50.60 & 72.09 & 57.27 & 47.57 & 61.10 \\
Ops-MM-embedding-v1-7B  & 8.3B & 59.30 & 78.80 & 49.40 & 67.16 & 46.12 & 43.95 & 57.45 \\
Qwen3-VL-Embedding-2B   & 2.1B & 63.60 & 83.50 & 42.50 & 65.67 & 41.33 & 37.74 & 55.72 \\
RzenEmbed-v1-7B         & 8.3B & 56.70 & 73.10 & 45.10 & 64.93 & 41.24 & 38.57 & 53.27 \\
VLM2Vec-V2.0            & 2.2B & 35.20 & 57.30 & 28.80 & 43.73 & 29.98 & 23.83 & 36.47 \\

\midrule
\rowcolor{OursBlue}
\multicolumn{9}{@{}l}{\textit{Omni embedders}}\\
\rowcolor{OursBlue}
LCO-Embedding-Omni-3B   & 4.7B & 39.00 & \underline{65.10} & 38.40 & 58.06 & 31.47 & 31.13 & 43.86 \\
\rowcolor{OursBlue}
LCO-Embedding-Omni-7B   & 8.9B & 30.50 & 64.50 & \underline{41.50} & 59.10 & 32.37 & 32.05 & 43.34 \\
\rowcolor{OursBlue}
omni-embed-nemotron-3B  & 4.7B & 22.40 & 36.60 & 35.70 & 55.22 & 32.37 & 27.09 & 34.90 \\
\rowcolor{OursBlue}
BidirLM-Omni-2.5B$^\dagger$ & 2.5B & 34.10 & 55.70 & 36.10 & 58.96 & 34.06 & 28.43 & 41.23 \\
\rowcolor{OursBlue}
e5-omni-3B              & 4.7B & 19.60 & 19.50 & 36.10 & 58.21 & \underline{36.35} & 27.76 & 32.92 \\
\rowcolor{OursBlue}
e5-omni-7B              & 8.9B & \underline{44.80} & 58.30 & 38.60 & \underline{59.25} & 35.16 & \underline{32.29} & \underline{44.73} \\
\rowcolor{OursBlue}
gemini-embedding-2$^\ddagger$ & ? & 32.20 & 56.30 & 47.50 & 66.12 & 45.82 & 40.89 & 48.14 \\
\rowcolor{BestGreen}
\multicolumn{9}{@{}l}{\textit{Ours: Omni-Embed-Mini}}\\
\rowcolor{BestGreen}
\quad 0.9B, no mining             & 935M & 10.90 & 14.50 &  9.70 & 19.10 &  9.26 &  7.73 & 11.87 \\
\rowcolor{BestGreen}
\quad 0.9B, text mining           & 935M & 10.80 & 13.60 &  9.10 &  7.76 &  4.28 &  2.23 &  7.96 \\
\rowcolor{BestGreen}
\quad 0.9B, media mining          & 935M & 13.20 & 15.10 & 11.60 & 18.51 &  7.57 &  4.91 & 11.81 \\
\rowcolor{BestGreen}
\quad 0.9B, text + media          & 935M & 16.70 & 27.40 & 18.50 & 27.61 & 12.75 &  9.83 & 18.80 \\
\rowcolor{BestGreen}
\quad 0.9B, t+m, orig.\ captions  & 935M &  9.00 &  7.40 &  5.30 &  3.88 &  1.69 &  1.18 &  4.74 \\
\rowcolor{BestGreen}
\quad 2.3B, text + media          & 2.3B & 62.40 & 83.40 & 42.20 & 65.52 & 40.84 & 36.74 & 55.18 \\
\rowcolor{ReleaseGreen}
\textbf{Omni-Embed-Mini-0.9B}     & 935M & 16.20 & 29.30 & 15.80 & 26.42 & 12.55 & 10.63 & 18.48 \\
\rowcolor{ReleaseGreen}
\textbf{Omni-Embed-Mini-2.3B}     & 2.3B & \textbf{62.40} & \textbf{83.40} & \textbf{42.20} & \textbf{65.52} & \textbf{40.84} & \textbf{36.74} & \textbf{55.18} \\
\bottomrule
\end{tabular}}
\caption{MMEB-V2 video per-task hit@1 ($\times 100$). \textbf{Task abbreviations:} HMDB51 (HMDB), UCF101 (UCF), MSR-VTT (MSRVTT), MSVD, DiDeMo, VATEX. \textbf{Task families} appear below each task name: V-CLS = zero-shot action classification accuracy; V-T2V = text-to-video retrieval hit@1. Across all omni-style models, the benchmark's own 8 pre-extracted frames per video are embedded as images and mean-pooled, so every model receives identical inputs under identical frame-level aggregation; the protocol may nevertheless understate a model's native temporal capability. Bolding, colour and footnote conventions follow \S\ref{supp:protocols}. $^\dagger$BidirLM-Omni-2.5B does not declare a video modality, but it accepts up to 8 image frames, so we evaluate it under the same 8-frame protocol.}
\label{supp:tab:mmeb-video}
\end{table*}

\subsection{ViDoRe-V3 (7 English domains)}
\label{supp:vidore}

\textbf{What it is.} ViDoRe-V3~\cite{loison2026vidore} is a
visual-document retrieval benchmark over visually rich enterprise
documents (financial reports, HR handbooks, scientific papers,
regulatory filings, pharmaceutical and industrial manuals): given a
text query, retrieve the relevant page-image.

\textbf{Subset evaluated.} We evaluate the 7 English-source domains
(Finance, HR, Industrial, Pharmaceuticals, Computer Science, Energy,
Physics); the French-only Finance task is excluded.

\textbf{Metric.} nDCG@10 per task; Avg is the mean across the 7
domains. The public ViDoRe leaderboard headline reports nDCG@5; we
report nDCG@10 to match the convention used elsewhere in this paper.
Per-domain results \mbox{are reported in Table~\ref{supp:tab:vidore}.}

\begin{table*}[!tp]
\centering
\setlength{\tabcolsep}{4pt}
\renewcommand{\arraystretch}{1.05}
\resizebox{\textwidth}{!}{%
\begin{tabular}{@{}l r rrrrrrr r@{}}
\toprule
\textbf{Model} & \textbf{Params}
& \multicolumn{1}{c}{Fin}
& \multicolumn{1}{c}{HR}
& \multicolumn{1}{c}{Ind}
& \multicolumn{1}{c}{Phr}
& \multicolumn{1}{c}{CS}
& \multicolumn{1}{c}{En}
& \multicolumn{1}{c}{Phy}
& \textbf{Avg} \\
\midrule

\multicolumn{10}{@{}l}{\textit{Visual-document retrievers}}\\
colqwen3.5-4.5B-v3            & 4.6B & 68.69 & 66.09 & 59.35 & 67.31 & 79.87 & 66.16 & 50.08 & 65.36 \\
nomic-embed-multimodal-7b     & 7.8B & 59.43 & 58.27 & 48.60 & 63.44 & 71.47 & 56.96 & 44.70 & 57.55 \\
nomic-embed-multimodal-3b     & 3.8B & 57.84 & 54.88 & 46.63 & 60.35 & 70.29 & 54.96 & 45.17 & 55.73 \\

\addlinespace[0.35em]
\multicolumn{10}{@{}l}{\textit{Vision--language embedders}}\\
llama-nemotron-embed-vl-1b-v2 & 1.7B & 64.53 & 60.59 & 47.85 & 63.52 & 74.01 & 57.72 & 43.65 & 58.84 \\
Qwen3-VL-Embedding-2B         & 2.1B & 62.23 & 58.45 & 47.93 & 62.51 & 73.48 & 56.81 & 46.64 & 58.29 \\
google/siglip-so400m-p14-384  & 878M & 11.04 & 11.22 &  8.91 & 22.14 & 19.58 &  7.57 & 10.25 & 12.96 \\

\addlinespace[0.35em]
\multicolumn{10}{@{}l}{\textit{Text retrievers} (operate on OCR'd page text)}\\
embeddinggemma-300m           & 308M & 51.03 & 49.06 & 42.44 & 60.28 & 67.07 & 55.74 & 43.21 & 52.69 \\
BAAI/bge-large-en             & 335M & 38.71 & 45.71 & 39.39 & 55.81 & 64.44 & 30.48 & 32.58 & 43.87 \\

\midrule
\rowcolor{OursBlue}
\multicolumn{10}{@{}l}{\textit{Omni embedders}}\\
\rowcolor{OursBlue}
LCO-Embedding-Omni-7B         & 8.9B & 55.13 & 55.31 & 41.94 & 61.14 & 69.43 & 55.44 & 44.64 & 54.72 \\
\rowcolor{OursBlue}
BidirLM-Omni-2.5B             & 2.5B & 58.87 & 51.58 & 41.17 & 56.23 & 68.85 & 51.92 & 42.77 & 53.06 \\
\rowcolor{OursBlue}
LCO-Embedding-Omni-3B         & 4.7B & 54.40 & 52.06 & 39.37 & 58.99 & 67.38 & 53.86 & 44.31 & 52.91 \\
\rowcolor{OursBlue}
omni-embed-nemotron-3B        & 4.7B & \underline{63.98} & 58.69 & \underline{49.80} & \underline{64.07} & 73.57 & \underline{59.71} & \textbf{47.59} & \underline{59.63} \\
\rowcolor{OursBlue}
e5-omni-3B                    & 4.7B & 60.41 & \underline{59.91} & 49.76 & 63.36 & 71.91 & 59.29 & 46.01 & 58.66 \\
\rowcolor{OursBlue}
e5-omni-7B                    & 8.9B & \textbf{65.41} & \textbf{61.27} & \textbf{49.83} & \textbf{66.48} & \underline{73.84} & \textbf{63.29} & \underline{46.86} & \textbf{61.00} \\
\rowcolor{OursBlue}
gemini-embedding-2$^\ddagger$ & ? & 63.12 & 57.88 & 46.91 & 63.52 & 72.04 & 51.06 & 45.13 & 57.09 \\
\rowcolor{BestGreen}
\multicolumn{10}{@{}l}{\textit{Ours: Omni-Embed-Mini}}\\
\rowcolor{BestGreen}
\quad 0.9B, no mining             & 935M & 45.97 & 40.54 & 36.87 & 51.67 & 62.35 & 47.67 & 40.62 & 46.53 \\
\rowcolor{BestGreen}
\quad 0.9B, text mining           & 935M & 45.43 & 41.77 & 37.86 & 52.25 & 62.24 & 48.41 & 40.44 & 46.91 \\
\rowcolor{BestGreen}
\quad 0.9B, media mining          & 935M & 46.85 & 44.57 & 36.01 & 52.56 & 61.98 & 46.61 & 36.36 & 46.42 \\
\rowcolor{BestGreen}
\quad 0.9B, text + media          & 935M & 46.76 & 41.04 & 37.35 & 52.58 & 62.00 & 48.11 & 41.03 & 46.98 \\
\rowcolor{BestGreen}
\quad 0.9B, t+m, orig.\ captions  & 935M & 38.24 & 39.56 & 32.47 & 48.84 & 57.75 & 42.80 & 35.82 & 42.21 \\
\rowcolor{BestGreen}
\quad 2.3B, text + media          & 2.3B & 62.64 & 57.61 & 47.89 & 62.01 & 74.43 & 56.11 & 46.00 & 58.10 \\
\rowcolor{ReleaseGreen}
\textbf{Omni-Embed-Mini-0.9B}     & 935M & 44.69 & 41.22 & 37.94 & 53.17 & 61.69 & 49.19 & 40.57 & 46.92 \\
\rowcolor{ReleaseGreen}
\textbf{Omni-Embed-Mini-2.3B}     & 2.3B & 62.64 & 57.61 & 47.89 & 62.01 & \textbf{74.43} & 56.11 & 46.00 & 58.10 \\
\bottomrule
\end{tabular}}
\caption{ViDoRe-V3 per-domain scores. All seven tasks are text-to-page retrieval (T2Pg) over visual-document page images, scored by nDCG@10 ($\times 100$). \textbf{Domain abbreviations:} Finance (Fin), HR, Industrial (Ind), Pharmaceuticals (Phr), Computer Science (CS), Energy (En), Physics (Phy). Bolding, colour and footnote conventions follow \S\ref{supp:protocols}. The mining ablation rows show that visual-document retrieval is largely insensitive to which side hard negatives are drawn from: the four 0.9B mining configurations agree to within 0.6\,nDCG@10 (46.42 to 46.98). What does move this modality is caption quality: replacing the dense cascaded captions with the datasets' original captions costs 4.8\,nDCG@10 (46.98 $\rightarrow$ 42.21).}
\label{supp:tab:vidore}
\end{table*}

\section{Cascaded-caption data generation}
\label{supp:datagen}

The caption slice of our omni-modal corpus is the sole training
corpus for the contrastive objective. Every row is the output of a
single multimodal call to Qwen3-Omni-30B-A3B-Instruct at
temperature 0.7 with a 2048-token output budget. The model
receives the raw media plus a modality-specific system prompt that
interpolates the source dataset's ground-truth caption as a
``Reference caption'' grounding prior; a short user-facing prompt is
drawn at random from a per-modality bank and concatenated with the
appropriate media tag.

\subsection{Modality system prompts}
\label{supp:prompts}

\newcommand{\promptbox}[2]{%
\begin{center}
\fbox{\begin{minipage}{0.96\linewidth}
\textbf{System prompt: #1}\\[0.3em]
\small\itshape #2
\end{minipage}}
\end{center}}

\promptbox{Audio}{%
Describe only what you hear; no opinions, speculation, or filler.
Be precise and factual. Do not hallucinate sounds not present.
Listen to this audio. Reference caption: ``\{original\_text\}''.
Write a dense, semantically rich description. Cover: each distinct
sound and its characteristics (pitch, volume, texture); temporal
order (what happens first, next, last); acoustic environment
(indoor/outdoor, reverb, background); patterns or repetitions. Be
concise; every word should convey information about the audio. One
paragraph.}

\promptbox{Music (audio music subsets)}{%
Describe only what you hear. Listen to this music. Reference
caption: ``\{original\_text\}''. Write a dense description covering:
instruments and timbres; melody, harmony, key, chord progressions;
rhythm, tempo (BPM estimate), time signature; genre, energy,
dynamics; structure; vocals if present (gender, style, lyrics
gist). One paragraph.}

\promptbox{Speech}{%
Listen to this speech audio. Write a single flowing paragraph
covering: (1) the speaker (age group, gender, voice pitch/tone,
clarity, accent); (2) speaking style (pace, loudness, emotion);
(3) a transition into the transcription using a casual phrase
followed by the marker [TRANSCRIPTION]; (4) audio quality, recording
environment, background sounds. Use the literal text [TRANSCRIPTION];
do not write out the actual words spoken. The placeholder is
post-processed into a transcription span \mbox{around the source transcript.}}

\promptbox{Image}{%
Look at this image. Reference caption: ``\{original\_text\}''. Write
a dense, factual description covering: subjects and objects (what,
how many, attributes); spatial layout and composition; colors,
lighting, and style; any actions or interactions; any text visible
in the image (OCR). Be concrete; avoid speculation about meaning or
\mbox{backstory. One paragraph.}}

\promptbox{Video}{%
Watch this video. Reference caption: ``\{original\_text\}''. Write a
dense description covering: subjects and setting; spatial layout;
the temporal sequence of events (what happens first, next, last);
camera motion or shot transitions; on-screen actions; any text or
captions visible \mbox{on screen. One paragraph.}}

\promptbox{Visual document}{%
You are an expert document analyst. Examine this document image and
provide a comprehensive description covering: document type and
purpose; overall layout and structure; all text content (headings,
body, labels); tables (column headers, row structure, key data);
charts and graphs (type, axes, trends); figures and diagrams; visual
formatting (colors, fonts, emphasis). Be thorough, \mbox{precise, and well-organised.}}

\subsection{User-prompt banks}
For each modality the user-side message is sampled from a small bank
of natural-language requests: e.g.\ ``what do you hear'', ``describe
this audio'', ``caption this'' for audio; ``what did they say'',
``transcribe this'' for speech; ``what do you see'', ``describe this
image'' for image; and ``what do you see'', ``describe this video'',
``what is happening in this video'' for video. Visual-document rows
use a single short user prompt; the grounding work is carried \mbox{by the system prompt.}

\subsection{Sample schema}
Each row of the caption slice carries a unique row identifier,
the media, a chat-style messages field holding the
user request and the model's dense caption, the modality label
(audio, speech, image, video, or visual-doc), the caption category
(always ``caption'' for this slice), the source dataset's
original short caption (used as the grounding prior), and metadata
fields for the source dataset name, sampling rate (16{,}000 for
audio/speech) and \mbox{frame rate (for video).}

\subsection{Qualitative dense-caption examples}
\label{supp:tab:cap-examples}

Table~\ref{supp:tab:cap-examples-tbl} shows one representative
example per modality from the caption slice. The middle column is a
brief written gloss of the underlying media (a one-line description
sitting between the original short caption and the dense caption);
the third column is the original short caption from the source
dataset (the grounding prior); the fourth column is the dense
caption produced by the cascaded captioner and used as the
contrastive teacher target.

\begin{table*}[!tp]
\centering
\footnotesize
\renewcommand{\arraystretch}{1.25}
\setlength{\tabcolsep}{4pt}
\begin{tabular}{@{}p{0.07\linewidth} p{0.22\linewidth} p{0.20\linewidth} p{0.46\linewidth}@{}}
\toprule
\textbf{Modality} & \textbf{Media (described)} & \textbf{Original caption} & \textbf{Dense caption} \\
\midrule
Audio &
A woman speaks in a public space while a slow heartbeat-like thump and faint ticking play in the background. &
A woman talks while a heartbeat and ticking play in the background. &
A woman's voice, speaking in a mid-to-low pitch, is present at the beginning of the clip, accompanied by faint, indistinct conversation from other people. Simultaneously, a deep, rhythmic thumping with a slow, steady tempo, resembling a heartbeat, begins and continues throughout. The background includes a persistent, low-level ambient hum and the general murmur of a public indoor space \mbox{with some reverberation.} \\
\midrule
Speech &
A young woman with a clear, neutral accent reads a short factual question in a calm, measured voice in a quiet indoor setting. &
What are organic plants understood to be? &
A young adult female with a high-pitched, clear, and articulate voice delivers the line in a standard, neutral accent. Her speaking style is measured and calm, with a moderate pace and consistent volume, conveying a neutral, informative affect. She says, ``What are organic plants understood to be?'' The audio is clean and crisp, likely recorded in a quiet indoor space with no discernible background \mbox{noise or reverberation.} \\
\midrule
Image &
\raisebox{-0.95\height}{\includegraphics[width=\linewidth]{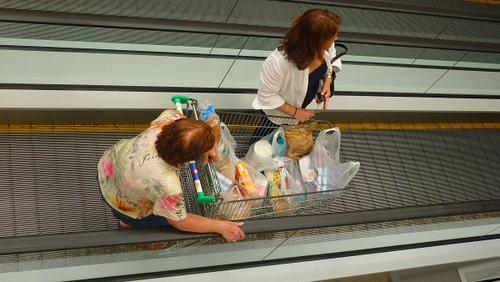}} &
Two women going down a conveyor belt with a cart full of groceries. &
The image, taken from a high-angle perspective, shows two women on a descending escalator. The woman in the foreground, wearing a floral-patterned short-sleeved shirt, leans forward, gripping the handle of a metal shopping cart. Behind her, another woman with brown hair, wearing a white top over a dark shirt, also holds the cart's handle. The cart is filled with various items, including a visible bag of potato chips and other grocery products. The setting appears to be indoors, with a neutral-toned, geometric-patterned wall in the background; lighting is even and diffuse, characteristic of an indoor public space. \\
\midrule
Video &
\begin{tabular}[t]{@{}c@{}}
  \includegraphics[width=0.95\linewidth]{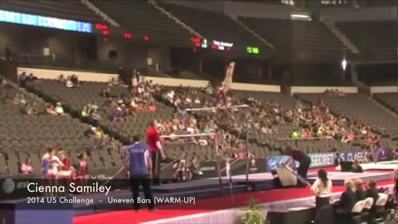}\\[1pt]
  \includegraphics[width=0.95\linewidth]{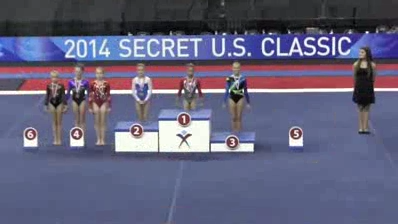}\\[1pt]
  \includegraphics[width=0.95\linewidth]{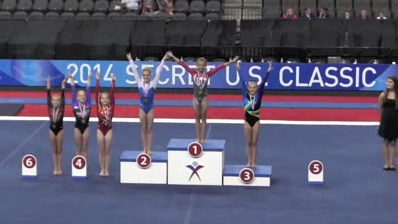}
\end{tabular} &
There is a girl gymnast who is \mbox{performing on the bars.} &
A gymnast, Cienna Samiley, performs on the uneven bars during the 2014 US Challenge warm-up. The scene is a large, indoor arena with sparse audience seating. The camera cuts to a podium where seven young female gymnasts stand, wearing leotards and medals, in front of a banner reading ``2014 SECRET U.S. CLASSIC''. A scoreboard graphic is superimposed, displaying the top six scores. The gymnasts on the podium then raise their arms and cheer. \\
\midrule
Visual doc &
\raisebox{-0.95\height}{\includegraphics[width=\linewidth]{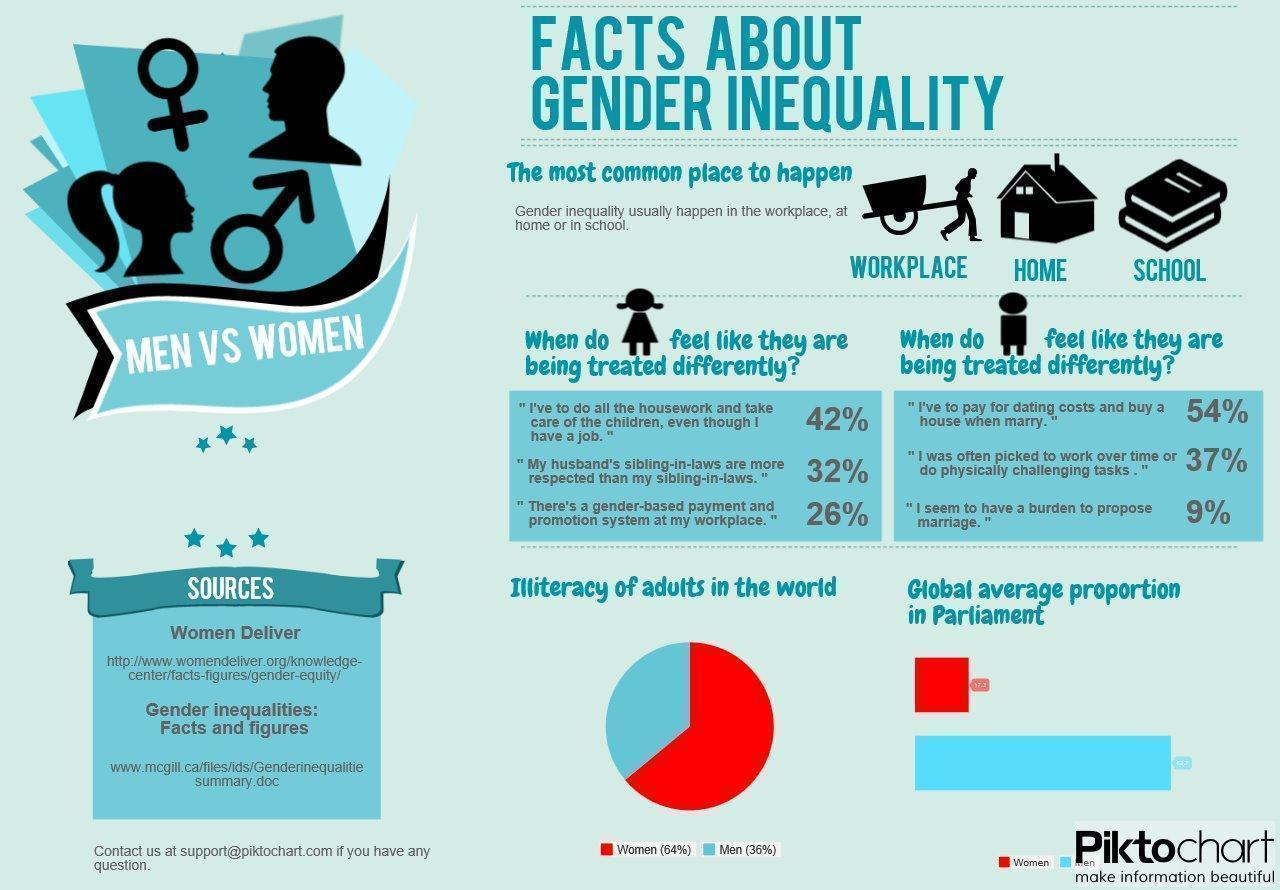}} &
(not available) &
A detailed analysis of an infographic titled ``Facts About Gender Inequality,'' which uses a combination of text, icons, and data visualizations to present information on gender disparities. The infographic is set against a light teal background, structured into several sections, with a bold sans-serif title at the top \dots\ Icons use simple black silhouettes (gender symbols, a wheelbarrow for the workplace, a house for home, books for school); the colour scheme is teal/red/black; data is presented via a pie chart and a bar chart with a consistent \mbox{red-for-women, blue-for-men encoding.} \\
\bottomrule
\end{tabular}
\caption{One sample per modality from the caption slice. Visual
documents in the source corpus carry no human-written original
caption; the cascaded captioner generates the dense description
directly from the page image, and we mark such cells
``(not available)''. Visual-document dense captions are themselves
substantially longer and richer than the other modalities; the
example shown here is truncated for layout (indicated by ``\dots'').
In the video example, the gymnast's name and the event are read from
on-screen text in the video itself, not inferred by the captioner.}
\label{supp:tab:cap-examples-tbl}
\end{table*}

\section{Per-source datasets, licenses, citations}
\label{supp:datasources}

The caption slice draws from 16 unique source datasets across 5
modalities, totalling 242{,}080 caption rows and 54.9 days of media.
Our generated dense captions are released under CC BY-SA 4.0; the
source media and their original annotations remain under the
licenses of their respective datasets, summarised in
Table~\ref{supp:tab:datasources}.

\subsection{Train/test overlap audit}
\label{supp:overlap}

Three of our training sources share a parent corpus with MMEB-V2
evaluation tasks, so we ran an identifier-level audit rather than
relying on split names. For each training sample we recovered the
original source identifier and compared it one to one against the
identifiers of the benchmark test samples. \textbf{The audit finds
zero shared items \mbox{in all three cases.}}

\textbf{MSR-VTT.} Evaluation uses the benchmark's \texttt{test\_1k}
split. Training and evaluation identifiers are disjoint and their
union is exactly the 10{,}000 videos of the corpus, so the two sides
are the complementary 9k/1k halves.

\textbf{ActivityNet.} Evaluation uses the \texttt{test} split,
614 unique videos after normalising YouTube identifiers; the training
side contributes 10{,}009 videos, of which 9{,}793 carry dense
captions. The intersection is empty on both the full training set and
the captioned subset. ActivityNet is present in training but is not
among the MMEB-V2 tasks we report; we audit it so that it is
confirmed clean \mbox{for future evaluation.}

\textbf{MSCOCO, via Recap-COCO-30K.} Matching is on the COCO
\texttt{image\_id} parsed from the image filenames, with the
evaluation images drawn from \texttt{val2014}. No training image
identifier appears among the MSCOCO-I2T or MSCOCO-T2I test
identifiers. We state this at the identifier level because that is
what we verified; we make no claim about how the upstream
Recap-COCO-30K release partitions the original COCO splits.

Because no overlapping items exist, no rows were removed and the
reported cross-modal numbers stand unchanged.

\begin{table*}[!tp]
\centering
\scriptsize
\setlength{\tabcolsep}{3pt}
\resizebox{\textwidth}{!}{%
\begin{tabular}{llrl}
\toprule
\textbf{Source} & \textbf{Modality} & \textbf{Samples} & \textbf{License} \\
\midrule
TED-LIUM 3~\cite{hernandez2018ted}                & speech     & 26{,}826 & CC BY-NC-ND 3.0 \\
HeySQuAD~\cite{wu2023heysquad}                    & speech     &  7{,}199 & CC BY 4.0 (HF declared); SQuAD text CC BY-SA 4.0 \\
Spoken Alpaca~\cite{shih2023gsqa}                 & speech     &  5{,}135 & CC BY-NC 4.0 \\
Gemini Speech~\cite{sb_2025}                      & speech     &  4{,}726 & Apache 2.0 \\
LibriSpeech~\cite{panayotov2015librispeech}       & speech     &  2{,}854 & CC BY 4.0 \\
WavCaps (FreeSound)~\cite{mei2024wavcaps}         & audio      & 26{,}217 & Academic / research only \\
WavCaps (AudioSet-SL)~\cite{mei2024wavcaps}       & audio      & 10{,}831 & Academic / research only \\
AudioCaps~\cite{kim2019audiocaps}                 & audio      &  4{,}518 & MIT (academic only) \\
WavCaps (BBC)~\cite{mei2024wavcaps}               & audio      &  3{,}115 & Academic / research only \\
Clotho~\cite{drossos2020clotho}                   & audio      &     394  & Tampere Univ.\ (non-commercial); audio: per-clip Freesound CC \\
WavCaps (SoundBible)~\cite{mei2024wavcaps}        & audio      &     118  & Academic / research only \\
ShareGPT-4o (image)~\cite{chen2024far}            & image      & 42{,}271 & MIT (academic only) \\
Recap-COCO-30K~\cite{li2024if}                    & image      & 22{,}474 & CC BY 4.0 \\
Flickr30k~\cite{young2014image}                   & image      & 21{,}398 & Research-only (Flickr ToU) \\
ActivityNet~\cite{krishna2017dense}               & video      & 27{,}526 & No explicit license (devkit MIT); YouTube ToS \\
MSR-VTT~\cite{xu2016msr}                          & video      &  6{,}612 & Research-only (Microsoft) \\
ShareGPT-4o (video)~\cite{chen2024far}            & video      &  1{,}555 & MIT (academic only) \\
VisRAG~\cite{yu2025visrag}                        & visual doc & 14{,}995 & Mixed (per sub-source) \\
ColPali~\cite{faysse2025colpali}                  & visual doc & 13{,}316 & Mixed (per sub-source) \\
\bottomrule
\end{tabular}}
\caption{Per-source breakdown of the caption slice of
our omni-modal corpus, with citations and source licenses. Our
generated captions are released under CC BY-SA 4.0; source media
keep the licenses listed here.}
\label{supp:tab:datasources}
\end{table*}
\section{Mining subsystem}
\label{supp:miner}

The miner runs as a single background thread that shares the live
training model with the main loop, avoiding the roughly 16\,GB extra
HBM cost of a separate model copy. Within-cycle weight drift between
miner and main thread is empirically negligible because the LoRA
delta per optimiser step is small and the embedding geometry is
dominated by the frozen backbone.

\paragraph{Cadence.} Every modality receives mined hard negatives.
The miner alternates between two cycle types, gated on the training
step counter. A \emph{text} cycle re-embeds the captions of all five
media modalities, builds a CPU FAISS index, and writes hard-negative
ids into the negative cache; it fires roughly ten times per epoch and
finishes in about thirty seconds. A \emph{media} cycle re-embeds the
samples of each media modality (speech, audio, image, video and
visual document) through the full model into per-modality indices;
it fires roughly five times per epoch and takes 12 to 20 minutes.
Triggers are coalesced, so multiple firings during a long cycle
collapse to a single follow-up.

\paragraph{Concurrency and cache semantics.} A single forward-lock
serialises miner and main-thread forward passes at batch granularity:
the main thread holds the lock for the entire forward, backward and
optimiser step, and the miner acquires it per encode-batch, yielding
fine-grained alternation rather than overlap. The negative cache is
keyed by sample id and overwrites on write, so media-mined entries
supersede earlier text-mined ones for the same sample. If the cache
is empty for a given sample at batch time the collator falls back to
in-batch negatives plus distillation, recovering the no-mining
ablation behaviour with no special-case code.

\section{Caption grounding and noise safeguards}
\label{supp:safeguards}

Because every teacher target is machine-generated, caption noise is a
direct risk to the learned geometry. Four properties of the recipe
limit how far that noise can propagate. First, the captioner is
\emph{grounded}: the source dataset's own caption is interpolated
into the prompt, so the model elaborates on a human-written anchor
rather than describing the media unconstrained. Second, the target is
a \emph{pooled embedding}, not the caption tokens; the student fits a
single vector, so local errors in wording are attenuated by pooling
rather than supervised token by token. Compression is not correction,
however: systematic hallucinations, omissions and semantic biases can
still propagate into the student through the pooled target. Third,
hard-negative mining applies a $0.92$ cosine cutoff, which discards
candidates too close to the anchor to be genuine negatives and so
removes the mining pairs most likely to be caption artefacts. Fourth,
the text path is frozen, so no amount of captioner bias can reach text
retrieval: whatever the captions contain, the text branch is the stock
backbone.

Visual documents are the informative case: they are the one modality
captioned with no reference caption to ground against, and they still
reach strong retrieval, which shows that fully generated captions can
provide effective supervision in this setting. It does not measure
robustness to hallucinated or systematically biased captions. These
are containment properties, not verification; we do not filter
captions for factuality, and the Limitations section states what that
leaves unmeasured.

\section{Additional ablations}
\label{supp:extra-ablations}

\paragraph{Matryoshka truncation.}

\begin{figure}[!htbp]
\centering
\includegraphics[width=\columnwidth]{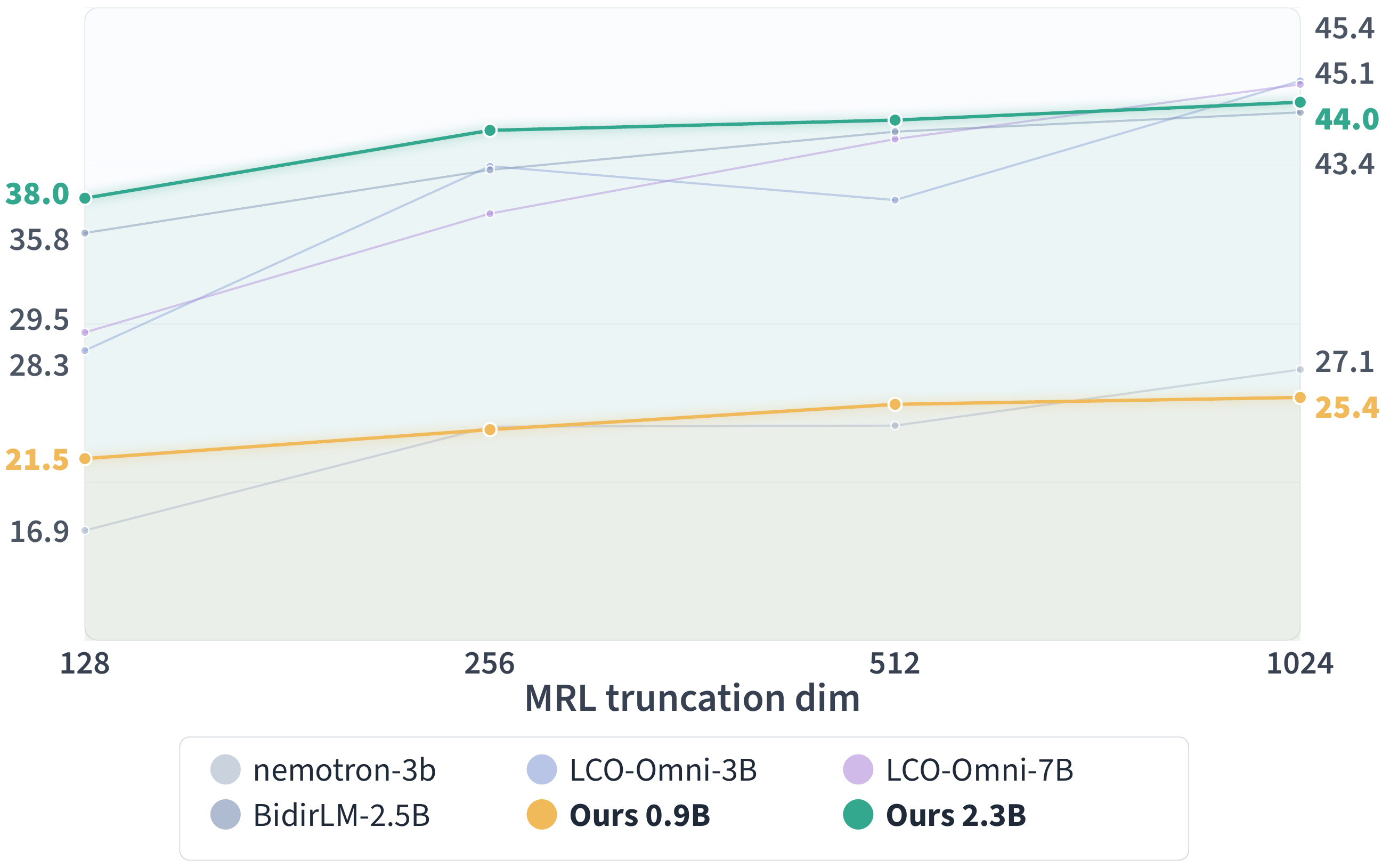}
\caption{\textbf{MRL truncation behaviour.} AVG over per-modality
probes (two retrieval tasks each from MTEB-v2, MAEB, MMEB-V2,
ViDoRe-V3) against truncation dimension
$d \in \{128, 256, 512, 1024\}$. Both \emph{Ours} curves are
near-monotone and stay close to full-dimension performance even at the
most aggressive 128-dim slice, whereas the omni-specific baselines
fall away far more steeply and their per-modality curves are not
monotone. Vision-language embedders that ship their own Matryoshka
training degrade comparably gently.}
\label{fig:mrl}
\end{figure}

Fig.~\ref{fig:mrl} plots the truncation curves summarised in
\S\ref{sec:ablations:mrl}. The additional analyses reported below are
a SciFact text-vs-speech NDCG@10 convergence probe, emergent
media-to-media alignment via the shared text axis, and the
single-stage data-loader with its phased optimiser schedule.

\paragraph{Seed sensitivity, per seed.}
Table~\ref{supp:tab:seeds} gives the per-seed scores behind the
summary in \S\ref{sec:ablations:mining}. All three runs use the
10-epoch no-mining configuration of the 0.9B variant and differ only
in the random seed.

\begin{table}[!htbp]
\centering
\small
\setlength{\tabcolsep}{4pt}
\begin{tabular}{@{}lrrrrr@{}}
\toprule
\textbf{Modality} & \textbf{s42} & \textbf{s71} & \textbf{s1234} & \textbf{Mean} & \textbf{SD} \\
\midrule
Text          & 49.57 & 49.57 & 49.57 & 49.57 & 0.00 \\
Speech        & 43.93 & 41.95 & 43.41 & 43.10 & 1.03 \\
Audio         & 33.28 & 34.22 & 34.37 & 33.96 & 0.59 \\
Image         & 27.96 & 27.35 & 27.13 & 27.48 & 0.43 \\
Video         & 20.86 & 19.61 & 18.99 & 19.82 & 0.95 \\
Visual Doc    & 46.76 & 46.21 & 43.47 & 45.48 & 1.76 \\
\bottomrule
\end{tabular}
\caption{\textbf{Seed sensitivity} of the 0.9B variant across seeds
42, 71 and 1234, on the 10-epoch no-mining configuration. Metrics
match the main tables. SD is the sample standard deviation over the
three seeds; the text row is identical by construction.}
\label{supp:tab:seeds}
\end{table}

\paragraph{Cross-modal alignment probe.}
During training we monitor a SciFact text-vs-speech NDCG@10
convergence probe every 10\% of an epoch. The synthesised speech
version of each claim is encoded through the full multimodal path
(Whisper + Dasheng $\to$ projectors $\to$ backbone) while the corpus
remains text. The two NDCG@10 curves converging is the headline
indicator that the audio path has reached the backbone's text
geometry.

\paragraph{Media-to-media alignment.}
\label{supp:emergent}
Routing every modality through the \emph{same} shared backbone and
aligning each one to the caption text places all media in one
geometry after training. We never train a media-to-media objective,
but because every modality shares the text axis as a common
reference frame, pairings such as image$\leftrightarrow$audio,
audio$\leftrightarrow$video and image$\leftrightarrow$document can
in principle be compared directly. This mirrors the emergent
alignment reported by ImageBind~\cite{girdhar2023imagebind}, here
induced by a text anchor rather than by contrasting modalities
pair-wise, with the frozen text-embedding backbone supplying the
shared metric.

\paragraph{Single-stage training.}
LLaVA~\cite{liu2023visual} and most VL models train in two distinct
stages with two distinct datasets. Omni-Embed-Mini runs as a
\emph{single-stage} pipeline: the dataset is loaded once and the
only ``phasing'' is internal to the optimisation schedule
(projectors only, then projectors plus encoder LoRA;
\S\ref{sec:method:design}).
The two phases share the same data-loader, the same dense cascaded
captions, and the same loss, with no hand-off. This is possible
because dense cascaded captions carry both alignment signal and task
signal in one row.

\section{Loss equations}
\label{supp:loss}

We give the full form of the Matryoshka SigLIP contrastive loss
summarised in Sec.~\ref{sec:method:loss}. Let $\hat a_i^{(d)}$,
$\hat p_j^{(d)}$, $\hat h_{ik}^{(d)}$ denote the L2-normalised,
$d$-truncated student, in-batch positive, and hard-negative
embeddings, the last being the $k$-th negative mined for anchor $i$.
The in-batch positive is the teacher embedding, so
$\hat a_i \equiv \hat s_i$ and $\hat p_i \equiv \hat t_i$.
With learned temperature $\tau=\exp(\log\tau)$
(clamped to $[0.01,0.3]$) and learned bias $b$ initialised to $-10$,
both shared across $d$,
{\small
\begin{align}
z^{(d)}_{ij} &= \tfrac{1}{\tau}\,\hat a_i^{(d)}\!\cdot\!\hat p_j^{(d)} + b, \quad
y_{ij}=\!\begin{cases} +1 & i=j\\ -1 & i\neq j\end{cases}\;,\\
\zeta^{(d)}_{ik} &= \tfrac{1}{\tau}\,\hat a_i^{(d)}\!\cdot\!\hat h_{ik}^{(d)} + b,
\end{align}
}
and the per-dim SigLIP loss is the mean over all concatenated pairs:
{\small
\begin{equation}
\mathcal{L}_{\text{siglip}}^{(d)} = -\frac{1}{B(B{+}K)}\!\left[\sum_{i,j}\!\log\sigma(y_{ij} z^{(d)}_{ij}) + \sum_{i,k}\!\log\sigma(-\zeta^{(d)}_{ik})\right]
\end{equation}
}
The Matryoshka aggregation uses $1/\sqrt{d}$ weighting,
$w_d = (1/\sqrt{d}) / \sum_{d'} (1/\sqrt{d'})$, upweighting
aggressive truncations to reflect their higher per-coordinate
retrieval cost. The cosine distillation term is
$\mathcal{L}_{\text{dist}} = \sum_d w_d \mathbb{E}_i [1 - \hat s_i^{(d)} \!\cdot\! \hat t_i^{(d)}]$,
and the total loss is
$\mathcal{L} = \mathcal{L}_{\text{contr}} + \lambda\,\mathcal{L}_{\text{dist}}$
with $\lambda=1$.

\section{Hyperparameters and \mbox{training schedule}}
\label{supp:hparams}

Table~\ref{supp:tab:hparams} lists the hyperparameters of the
smallest variant, Omni-Embed-Mini-0.9B. Training runs on 8 GPUs with
DDP, BF16 mixed precision, AdamW, and a single cosine schedule with
warmup. Batch size is held near-uniform across modalities (image,
visual document, audio and speech at 32; video at 8 due to its
patch-token footprint).

Every number we report for a given variant comes from a single
checkpoint of that variant, evaluated across all six modalities.

\begin{table}[!b]
\centering
\small
\setlength{\tabcolsep}{4pt}
\begin{tabularx}{\columnwidth}{@{}>{\raggedright\arraybackslash}Xr@{}}
\toprule
\textbf{Component} & \textbf{Params} \\
\midrule
Frozen text backbone (Qwen3-Embedding-0.6B) & 595.78\,M \\
Vision encoder (ViT from Qwen3.5-0.8B) & 100.59\,M \\
Whisper-small encoder & 88.15\,M \\
Dasheng-base encoder & 85.46\,M \\
Projectors and pooler & 65.37\,M \\
\midrule
\textbf{Total at inference} & \textbf{935.35\,M} \\
\bottomrule
\end{tabularx}
\caption{\textbf{Per-component parameter breakdown} of
Omni-Embed-Mini-0.9B, expanding its total in
Table~\ref{tab:params}. Training updates the projectors and pooler
plus 2.95\,M of media-encoder LoRA, 68.32\,M in total. The 2.3B
variant replaces the backbone and its separate vision encoder with
Qwen3-VL-Embedding-2B and its native vision path (2126.99\,M), keeps
the same audio encoders, and uses larger projectors (139.61\,M),
\mbox{giving 2440.21\,M at inference.}}
\label{supp:tab:params-breakdown}
\end{table}

\begin{table*}[!tp]
\centering
\small
\setlength{\tabcolsep}{7pt}
\renewcommand{\arraystretch}{1.08}
\begin{tabularx}{\textwidth}{@{}lX@{}}
\toprule
\textbf{Component} & \textbf{Setting} \\
\midrule
\multicolumn{2}{@{}l}{\textit{Architecture}} \\
Backbone, frozen & Qwen3-Embedding-0.6B \\
Hidden size & 1024 \\
MRL dimensions & $\{128, 256, 512, 1024\}$ \\
Vision encoder & ViT extracted from Qwen3.5-0.8B \\
Image resolution & $224 \times 224$ \\
Video tokens / max frames & 196 / 96 \\
Audio encoders, both variants & Whisper-small encoder + Dasheng-base \\
Tokens per audio encoder & 128 \\
Max audio duration & 60\,s \\
\midrule
\multicolumn{2}{@{}l}{\textit{Optimisation}} \\
LoRA rank / $\alpha$ / dropout & 16 / 32 / 0.0 \\
LR, audio adapters & $2 \times 10^{-4}$ \\
LR, vision adapters & $1 \times 10^{-4}$ \\
LR, encoder LoRA & $1 \times 10^{-4}$ \\
Optimizer & AdamW, weight decay 0.01 \\
Warmup & 2\% of total steps \\
Max gradient norm & 1.0 \\
Max sequence length & 2048 \\
Batch size per GPU, image / visual-doc / audio / speech / video & 32 / 32 / 32 / 32 / 8 \\
Epochs & 2, for the reported ablation runs \\
Mixed precision & BF16 \\
GPUs & 8, DDP \\
\midrule
\multicolumn{2}{@{}l}{\textit{Loss and schedule}} \\
Distillation loss weight & 1.0, cosine \\
Initial temperature & 0.07, learnable \\
Phase-2 switch & 20\% of total steps \\
\midrule
\multicolumn{2}{@{}l}{\textit{Hard-negative mining}} \\
Text-mining refresh & $10\times$ per epoch \\
Media-mining refresh & $5\times$ per epoch \\
Text index size & 50{,}000 captions \\
Media index size & 10{,}000 per modality \\
Hard negatives per sample & 5 \\
False-negative threshold & 0.92, cosine \\
\bottomrule
\end{tabularx}
\caption{\textbf{Hyperparameters} for the Omni-Embed-Mini-0.9B variant. The 2.3B
variant differs in backbone, using Qwen3-VL-Embedding-2B with native
vision, and keeps the same Whisper-small and Dasheng-base audio
encoders. Proportional batch sizes are adjusted, while optimisation
settings remain unchanged.}
\label{supp:tab:hparams}
\end{table*}

\section{Additional Limitations}
\label{supp:limits}

\begin{itemize}[leftmargin=*,itemsep=2pt,topsep=2pt]
\item \textbf{Vision modes diverge between the 0.9B and 2.3B
variants.} The 0.9B variant uses a ViT extracted from the
vision-native Qwen3.5-0.8B model because the 0.6B text backbone has
no vision module; the 2.3B variant
reuses its backbone's native vision module. The training
configuration accommodates both, but cross-variant ablations of the
two vision strategies are \mbox{left to future work.}
\item \textbf{Visual-document captions are roughly six times longer
than other modalities.} Visual-document dense captions average about
4{,}157 characters versus about 582 for audio. The frozen backbone's
2{,}048-token context truncates the long tail of visual-document
captions; truncation is consistent between teacher and student paths
so cosine alignment is unaffected, but the upper end of visual-doc
detail is lost from the contrastive signal.
\item \textbf{Video uses per-frame mean-pooled embeddings on
MMEB-V2.} For all omni-style models, the benchmark's own 8 frames
per video are embedded as images and mean-pooled. This keeps results
directly comparable but caps video performance below what a native
\mbox{temporal encoder could reach.}
\end{itemize}